\documentclass[9.5pt,journal,final,finalsubmission,twocolumn]{IEEEtran}
\usepackage{amsfonts}
\usepackage{stmaryrd}
\usepackage{mathrsfs}
\usepackage{amssymb}
\usepackage{cite}
\usepackage{algorithmic}
\usepackage{graphicx}
\usepackage{psfrag}
\usepackage{amsmath}
\usepackage{txfonts}
\usepackage{latexsym,bm}
\newtheorem{algorithm}{Algorithm}[section]
\newtheorem{theorem}{Theorem}[section]

\newcommand{\R}{{\mathbb R}}

\newcommand{\ds}{\displaystyle}

\newcommand{\be}{\begin{eqnarray}}
\newcommand{\ben}{\begin{eqnarray*}}
\newcommand{\en}{\end{eqnarray}}
\newcommand{\enn}{\end{eqnarray*}}

\begin{document}
\title{Isometric Multi-Manifolds Learning}
\author{Mingyu Fan\thanks{M. Fan and B. Zhang are with LSEC and the Institute of Applied Mathematics,
AMSS, Chinese Academy of Sciences, Beijing 100190, China ~(email:
fanmingyu@amss.ac.cn, b.zhang@amt.ac.cn)}, Hong Qiao\thanks{H. Qiao
is with the Institute of Automation, Chinese Academy of Sciences,
Beijing 100190, China (email: hong.qiao@mail.ia.ac.cn)}, {\it Senior
Member, IEEE,} and Bo Zhang\thanks{\bf Corresponding author: Bo Zhang}
}

\maketitle

\begin{abstract}
Isometric feature mapping (Isomap) is a promising manifold learning
method. However, Isomap fails to work on data which distribute on
clusters in a single manifold or manifolds. Many works have been
done on extending Isomap to multi-manifolds learning. In this paper,
we first proposed a new multi-manifolds learning algorithm (M-Isomap) with
help of a general procedure. The new algorithm preserves
intra-manifold geodesics and multiple inter-manifolds edges
precisely. Compared with previous methods, this algorithm can
isometrically learn data distributed on several manifolds. Secondly,
the original multi-cluster manifold learning algorithm first proposed in \cite{DCIsomap}
and called D-C Isomap has been revised so that the revised D-C Isomap can learn multi-manifolds data.
Finally, the features and effectiveness of the proposed multi-manifolds learning
algorithms are demonstrated and compared through experiments.
\end{abstract}


\section{Introduction}

Challenges, known as "the curse of dimensionality", are usually confronted when scientists
are dealing with high dimensional data. Dimensionality reduction is a promising tool to circumvent
these problems. Principal component analysis (PCA) \cite{PCA} and
multidimensional scaling (MDS) \cite{MDS} are two important linear
dimensionality reduction methods. Due to their linear model
assumption, both of the methods will fail to discover nonlinear
intrinsic structures of data.

Recently, there are more and more interests in nonlinear dimensionality reduction (NLDR).
{\bf NLDR is used to learn
nonlinear intrinsic structures of data, which is considered to be
the first step of machine learning and pattern recognition
\cite{ScienceMachLearning}. }
Two interesting nonlinear
dimensionality reduction methods based on the notion of manifold
learning \cite{manifoldPerception}, isometric feature mapping
(Isomap) \cite{ISOMAP2000} and local linear embedding (LLE)
\cite{LLE2000}, have been introduced in {\em SCIENCE 2000}. LLE
assumes that data points locally distribute on a linear patch of a
manifold. It preserves local linear coefficients, which best
reconstruct each data point by its neighbors, into a lower
dimensional space. Isomap is based on the classical MDS method.
Instead of preserving pairwise Euclidean distance, it preserves the
geodesic distance on the manifold. The geodesic between two data
points is approximated by the shortest path on a constructed graph.
Both of the methods are computationally efficient and able to
achieve global optimality. There are also many other important
nonlinear dimensionality reduction methods. Laplacian eigenmap
\cite{laplacianEigenmap} utilizes an approximation of the
Laplace-Beltrami operator on manifolds to provide an optimal
embedding. Hessian LLE \cite{hessianLLE} resembles Laplacian
eigenmap by using an approximation of the Hessian operator instead
of Laplacian operator. The local tangent space alignment (LTSA)
\cite{LTSA} method learns the local geometry by constructing a local
tangent space of each data point and then aligns these local tangent
spaces into a single global coordinates system with respect to the
underlying manifold. Diffusion maps \cite{DiffusionMaps} applies
diffusion semigroups to produce multi-scale geometries to represent
complex structures. Riemannian manifold learning (RML) \cite{RML}
method uses the constructed Riemannian normal coordinate chart to
map the input data into a lower dimensional space. NLDR is a fast
growing research activity and has been proved very useful in many
fields and applications, such as classification using Isomap
\cite{Classify-Isomap} and Laplacian eigenmap
\cite{Classify-laplacian}, geometry based semi-supervised learning
method using Laplacian eigenmap \cite{Semi-learning}, data
visualization \cite{NDRforVisual&Classify} and time series manifold
learning \cite{STisomap,IsomapTimeSeries}.

As Isomap emphasizes on the global geometric relationship of data points, it is very illustrative in data
visualization and pattern analysis \cite{incrementaIsomap}. Although Isomap algorithm
implicitly requires the data set to be convex \cite{hessianLLE}, it still provides very meaningful results on non-convex data sets.
In this paper, we will focus on extending Isomap to multi-manifolds learning. {\bf The first  step of Isomap
algorithm is to construct a neighborhood graph which connects all the data points.}
This step is of vital importance because the success of the following steps depends on how well the
constructed neighborhood graph is. However, it is hard to build a totally connected neighborhood graph
in order to guarantee the topological stability of the classical Isomap algorithm when points of the data set
distribute on clusters in a manifold or manifolds (multiple manifolds).
It should be remarked that several methods have been proposed to extend Isomap to multi-manifolds data,
and some of them are based on providing new neighborhood graph construction algorithms.
For example, Wu et al \cite{ExtendedIsomaplearning} introduced a split and augment procedure for
neighborhood graph construction which could produce a totally connected neighborhood graph.
In a series of papers \cite{kEC2005,kMST2004,KVC2006,MinkST2005}, Yang introduced several neighborhood graph
construction algorithms using techniques from discrete mathematics, graph theory. Meng et al \cite{DCIsomap}
proposed a decomposition and composition Isomap (D-C Isomap).

The rest of the paper is organized as follows. In Section \ref{sec2}, the main issues and limitations of
the classical Isomap algorithm are presented. The problem of multi-manifolds learning is also investigated.
In Section \ref{sec3}, previous methods on multi-manifolds learning are briefly introduced and discussed.
In Section \ref{sec4}, a general procedure for designing multi-manifolds learning algorithms is first
proposed. With the proposed procedure, a new multi-manifolds learning algorithm (M-Isomap) is then designed
and analyzed. In addition, the original D-C Isomap algorithm is revised to overcome its main limitation.
In Section \ref{sec5}, the effectiveness of these multi-manifolds learning algorithms has been demonstrated by
experiments. Comparisons of these algorithms have also been made. Some concluding remarks are provided in
Section \ref{sec6}.

\section{Classical Isometric Feature Mapping and Its Limitations}\label{sec2}

Isomap is an efficient NLDR algorithm to recover the intrinsic geometric structure of a data set
if the data points lie on a single manifold \cite{ISOMAP2000}.
Assume that the data set $X=\{x_1,x_2,\cdots,x_N\}$ is in a high dimensional space $\R^D$ and
the feature space is $\R^d$. Then the classical Isomap algorithm has the following three steps \cite{ISOMAP2000}.

\begin{enumerate}

\item[]  Step 1: Identify the neighbors of all the data points to construct
a neighborhood graph. With the given parameter $k$ or $\epsilon$,
there are two ways to construct a neighborhood graph for $X$:

\begin{itemize}
\item if $x_j$ is one of $x_i$'s $k$ nearest neighbors, they are connected by an edge (the k-NN method).

\item if $x_i$ and $x_j$ satisfy $\|x_i-x_j\|<\varepsilon$, they are connected
by an edge (the $\varepsilon$-NN method).

\end{itemize}

\item[] Step 2: Use Dijkstra's or Floyd-Warshall's algorithm to compute the length of the shortest path
$d_G(x_i,x_j)$ between any two data points $x_i$ and $x_j$ on the graph. It is proved that
$d_G(x_i,x_j)$ is a good approximation of the geodesic distance
$d_M(x_i,x_j)$ on the manifold as the number of data points increases.

\item[] Step 3:  Perform the classical MDS on the graph distance matrix $D_G$ whose $(i,j)$-th element is $d_G(i,j)$.
Minimize the cost function
\begin{equation*}
E(Y)=\|\tau(D_G)-\tau(D_Y)\|_{F_2}
\end{equation*}
The operator $\tau$ is defined as $\ds\tau(D)=-\frac12{HSH}$, where
$\ds H=I-\frac{1}{n}ee^T$ with $I$ the identity matrix and $e=(1,1,\cdots,1)^T$,
$S=(D_{ij}^2)$ with $D_{ij}$ being the $(i,j)$-th element of $D$ and $D_Y=(\|y_i-y_j\|)$.
Assume that, in descending order, $\lambda_i$ is the $i$-th eigenvalue of $\tau(D_G)$ with
the corresponding eigenvector $\nu_i$. Then the low-dimensional embedding $Y$ is given by:

\ben
Y=[y_1,,y_2,\cdots,y_n]=\begin{pmatrix} \sqrt{\lambda_1}\nu_1^T\\
                                        \cdots \\ \sqrt{\lambda_d}\nu_d^T
                         \end{pmatrix}
\enn

\end{enumerate}

The property of the Isomap algorithm is well understood \cite{proof,ContinuumIsomap}.
However, the success of the Isomap algorithm depends on two issues.
One is how to choose the correct intrinsic dimensionality $d$.
Setting a lower dimensionality $d$ will lead to a loss of data structure information.
On the other hand, setting a higher dimensionality $d$ will make some redundant information to be kept.
This issue has been well investigated so far.
The other issue is the quality of the constructed neighborhood graph.
It is known that constructing an appropriate neighborhood graph is still a tricky task.
Both the k-NN and $\varepsilon$-NN methods have their limitations.
Under the assumption that data points distribute on a single manifold, if the neighborhood size $k$ or
$\varepsilon$ is chosen to be too small, the constructed neighborhood graph will be very
sparse and therefore the geodesics can not be satisfactorily approximated. On the other hand,
if the neighborhood size $k$ or $\epsilon$ is chosen to be too large, short-circuit edges may occur
which will have a significant negative influence on the topological stability of the
Isomap algorithm \cite{IsomapTopoStability}.

Nonetheless, if data points distribute uniformly on one manifold, then both the "short circuit" problem
and the "discontinuity" problem can be circumvented by carefully choosing an appropriate neighborhood
size $k$ or $\varepsilon$. However, if data points distribute on several clusters or manifolds,
then neither of the k-NN method and the $\varepsilon$-NN method can guarantee that the whole data set is
totally connected and the geodesics is satisfactorily approximated.

{ \bf However, both data missing and data mixture are common problems in practical data analysis.}
These two cases often cause data points to distribute on different clusters in a manifold or manifolds.
Data points on different manifolds may have different input dimensionality $D$
(the dimensionality of the ambient space). This usually happens in the case of data mixture.
On the other hand, learning different data manifolds may need different values of input parameters, that is,
appropriate neighborhood size ($k$ or $\varepsilon$) and intrinsic dimensionality $d$ for each data manifold.
{\bf In this paper, we will focus on designing new multi-manifolds learning
algorithms for data distributing on multiple manifolds. The case when data points distribute on pieces of
a single manifold is referred to as multi-cluster manifold
learning, while the case when data points distribute on multiple manifolds is referred to as
multi-manifolds learning.}

\section{Previous Works on Multi-Manifolds Learning}\label{sec3}

\subsection{Multi-manifolds learning by new neighborhood graph construction method}

Wu and Chan \cite{ExtendedIsomaplearning} proposed a split-augment approach to construct
a neighborhood graph. Their method can be regarded as a variation of the $k$-NN method
and can be summarized as below:
\begin{enumerate}
\item [1.] the $k$-NN method is applied to the data set.
Every data point is connected with its neighbors. If the data lies
on multiple manifolds, several disconnected graph components (data manifolds) will be formed.
\item [2.] Each couple of graph components are connected by their
nearest couple of inter-components points.
\end{enumerate}

This method is simple to implement and has the same computational
complexity as the $k$-NN method has. However, as there is only one edge
connecting every two graph components, geodesics across components
are poorly approximated and, meanwhile, their low dimensional embedding
can be rotated arbitrarily. This method can not be directly applied
to data lying on three or more data manifolds. If more than two
graph components exist, the intra-component shortest distances may be
changed in the totally connected graph.

Yang \cite{kMST2004,MinkST2005,kEC2005,KVC2006} introduced four methods to construct a connected
neighborhood graph. The $k$ minimum spanning trees ($k$-MST) method \cite{kMST2004} repeatedly
extracts $k$ minimum spanning trees (MSTs) from the complete Euclidean graph of all data points.
Edges of the $k$ MSTs form a $k$-connected neighborhood graph. Instead of extracting $k$ MSTs,
the minimum-$k$-spanning trees (min-$k$-ST) method \cite{MinkST2005} finds $k$ edge-disjoint spanning
trees from the complete Euclidean graph, and the sum of the total edge length of the $k$ edge-disjoint
spanning trees attains a minimum. The $k$-edge-connected ($k$-EC) method \cite{kEC2005} constructs a
connected neighborhood graph by adding edges in a non-increasing
order from the complete Euclidean graph. An edge is added if its two
end vertices do not have $k$ edge-disjoint paths connected with each
other. The $k$-vertices-connected ($k$-VC) method \cite{KVC2006} adds
edges in a non-increasing order from the complete Euclidean graph, where
an edge is added if its two end vertices would be disconnected by removing some $k-1$ vertices.
And the constructed neighborhood graph would not be disconnected by removing any $k-1$ vertices.

The methods introduced in \cite{kMST2004,MinkST2005,kEC2005,KVC2006} have the following advantages
over the $k$-NN method. First, the local neighbor relationship is affected by the global distribution
of data points. This is beneficial for adaptively preservation of the global geometric metrics.
Secondly, these methods can guarantee that the constructed neighborhood graph is totally connected.
Compared with the $k$-NN method, Yang's methods construct a neighborhood graph with more
edges corresponding to the same neighborhood size $k$. This property ensures the quality of the neighborhood graphs.

\subsection{Multi-manifolds learning by decomposition-composition Isomap}

In \cite{DCIsomap}, Meng et al. proposed a decomposition-composition method (D-C Isomap)
which extends Isomap to multi-cluster manifold learning. The purpose of the method is
to preserve intra-cluster and inter-cluster distances separately.
In the next section, we will introduce a revised version of the D-C Isomap to extend
the application range of the original D-C Isomap. To this end, we present the details of
the D-C Isomap algorithm as follows.

\noindent {\bf Step I: Decomposition process}
\begin{enumerate}
\item [1.] {\bf Given an appropriate neighborhood size $k$ or $\varepsilon$, if data is comprised of multiple clusters,
several disconnected graph components will be formed when each data point is
connected with its neighbors by the $k$-NN or $\varepsilon$-NN method.
\item [2.] Assume that there are $M$ graph components and a graph component is also considered as a cluster.
Data points of the $m$-th cluster is denoted as $X^m=\{x_1^m,\cdots,x_{l_m}^m\}$. Clusters $X^m$ and $X^n$
are connected by their nearest
inter-cluster edge, whose ending vertices are assumed as $nx^m_n$ and $nx^n_m$ and edge length as $d^0_{m,n}$.
\item [3.] Apply the $k$-NN Isomap or $\varepsilon$-NN Isomap on cluster $X^m$.
Denote by $D^m=(D^m_{i,j})$ the geodesic distance matrix for $X^m$, by $Y^m=\{y^m_1,\cdots,y^m_{l_m}\}$
the corresponding low-dimensional embedding to $X^m$ and by $ny^m_n$ the embedding point
corresponding to $nx^m_n$, where $ny^m_n\in Y^m$.}
\end{enumerate}

\noindent {\bf Step II: Composition process}

\begin{enumerate}
\item [1.] The set of centers of clusters is denoted as $CX=\{cx_1,\cdots,cx_M\}$,
where each center is computed by
\ben\label{e31}
cx_m=\underset{x_i \in X^m}{\arg\min}\left(\underset{x_j\in X^m}{\max}(D_{ij}^m)\right)\qquad m=1,\cdots,M .
\enn

\item [2.] The distance matrix for $CX$ can be computed by
\ben\label{e32}
\tilde{D}=\{\tilde{D}_{mn}\},\qquad \tilde{D}_{mn}= \left \{
\begin{array}{ll}
d_{m,n}+d_{m,n}^0+d_{n,m} & m\not=n  \\ 0, & m=n
\end{array}\right.
\enn
where $d_{m,n}$ is the distance of the shortest path between $cx_m$ and
$nx_n^m$ on the graph component $X^m$.

\item [3.] Plug the distance matrix $\tilde{D}$ and the neighborhood size
$d$ into the classical Isomap algorithm. The embedding of $CX$ is denoted by
$CY=\{cy_1,\cdots,cy_M\} \subset \mathbb{R}^d$ ($CY$ is called the translation reference set).
Assume that the $d$ nearest neighbors of $cx_m$ are $\{cx_{m_1},\cdots,cx_{m_d}\}$.
Then the low-dimensional representation corresponding to $nx_{m_i}^m$ can be computed as
\ben\label{e33}
sy_{m_i}^m=cy_m+\frac{d_{m,m_i}}{d_{m,m_i}+d_{m,m_i}^0+d_{m_i,m}}(cy_{m_i}-cy_m),
\enn
where $i=1,\cdots,d$ and $m=1,\cdots,M.$

\item [4.] Construct the rotation matrix $\mathcal{A}_m$ for $Y^m$ with $m=1,\cdots,M.$
Assume that $QN_m$ is the principal component matrix of $NY_m=\{ny_{m1}^m,\cdots,ny_{m_d}^m\}$
and $QS_m$ is the principal component matrix of $SY_m=\{sy_{m1}^m,\cdots,sy_{m_d}^m\}$.
Then the rotation matrix for $Y^m$ is $\mathcal{A}_m=QS_m QN_m^T$.

\item [5.] Transform $Y^m$ ($m=1,\cdots,M$) into a single coordinate system{\bf by Euclidean} transformations:
\ben\label{e34}
FY^m=\{fy_i^m=\mathcal{A}_m y_i^m +cy_m, i=1,\cdots,l_m\},\quad m=1,\cdots,M.
\enn
Then $Y=\bigcup_{m=1}^M FY^m$ is the final output.
\end{enumerate}

First, the D-C Isomap algorithm reduces the dimensionality of clusters separately and meanwhile,
preserves a skeleton of the whole data.
Secondly, using the Euclidean transformations, the embedding of each cluster is placed
into the corresponding position by referring to the skeleton.
In this way, the intra-cluster geodesics are exactly preserved.
Since the D-C Isomap method uses circumcenters to construct the skeleton of the whole data,
its learning results depend on the mutual position of these circumcenters, which would make
the learning results unstable.
On the other hand, it is known that at least $d+1$ reference points are needed to anchor
a $d$-dimensional simplex. However, in the D-C Isomap algorithm, the number of the reference
data points is limited by the number of clusters.

\subsection{Constrained Maximum Variance Mapping}

Recently in \cite{CMVM}, Li et al. proposed the constrained maximum variance mapping (CMVM)
algorithm for multi-manifolds learning. The CMVM method is proposed based on the notion of
maximizing the dissimilarity between classes while holding up the intra-class similarity.

\section{Isometric Multi-Manifolds Learning}\label{sec4}

\begin{table*}
\caption{Symbols and Variables used in the Algorithms} \centering
\begin{tabular}{|ll|}
\hline
$X=\{x_i\}_{i=1}^N$ & The total data set, with $x_i \in \R^D$\\
$X^m=\{x_i^m\}_{i=1}^{S_m}$ & The m-th data manifold, where $m=1,\cdots, M$\\
$Y^m=\{y_i^m\}_{i=1}^{S_m}$ & Low dimensional embedding of $X^m$ \\
$D_{mn}=(d_G(x_i^m,x_j^n))$  &  Matrix of geodesic distances across data manifolds $X^m$ and $X^n$\\
$fx_n^m$, $fx_m^n$  & The furthest couple of data points in $ X^m$ and $X^n$, with $f_n^m\in X^m$ $f_m^n\in X^n$\\
$\{x_{n(i)}^m\}_{i=1}^k$ & Subset of $X^m$ whose elements are the nearest data points to $X^n$\\
$I^m=\{x_i^m\}_{i=1}^{l_m}$ &  The selected data points from $X^m$ to construct the skeleton $I$ \\
$Y_I^m$  &  Low-dimensional embedding of $I^m$ with $Y_I^m \subset Y^m$\\
$I=\bigcup_{m=1}^M I^m $   &  Points which are used to construct a skeleton of $X$\\
$D_I$   &  Approximated geodesic distance matrix for skeleton $I$\\
$RY_I$  & Low-dimensional embedding of the skeleton $I$ \\
$RY_I^m$ & Low-dimensional embedding of $I^m$ with $RY_I^m\subset RY_I$. $RY_I^m$ will also \\
                     & be referred to as the transformation reference for $Y^m$ \\
$nx_j^i$ & The point in $X^i$ which is the nearest to $X^j$ or the inter-cluster point \\
         & in the D-C Isomap algorithm.
\\\hline
\end{tabular}
\end{table*}

In this section, we first introduce a general procedure for the design of
isometric multi-manifolds learning algorithms and then present our new multi-manifolds
learning algorithm called M-Isomap. Finally, we make a revision of the original D-C Isomap
algorithm to extend its application range.

\subsection{The general procedure for isometric multi-manifolds learning}

Many previous methods extend Isomap to multi-manifolds learning by
revising the neighborhood graph construction step of the Isomap
algorithm \cite{ExtendedIsomaplearning,kEC2005,kMST2004,KVC2006,MinkST2005}.
However, the shortest paths across clusters or data manifolds are bad
approximations of geodesics. In Isomap, bad local approximation
always leads to deformation of the global low-dimensional embedding.

Assumed that $\Omega$ is an open, convex and compact set in $\R^d$ and
$f:\Omega\to\R^D$ is a continues mapping, where $d<<D$.
Then $f(\Omega)=\mathcal {M}$ is defined as a $d$ dimensional parameterized manifold.
Let $K(x,y)$ $(x,y\in\mathcal{M})$ be a specially defined kernel.
Then a reproducing kernel Hilbert space (RKHS) $\mathscr{H}$ is constructed with this kernel.
Denote by $\phiup_j(x)$ the eigenfunction corresponding to the $j$-th largest eigenvalue
$\lambda_j$ of $K(x,y)$ in $\mathscr{H}$, which is also the $j$-th element of the Isomap embedding.
The geodesic distance on the manifold $\mathcal {M}$ is written as
$$
d^2(x,y)=d^2(f(\tau),f(\overset{\wedge}{\tau}))
  = \alpha\|\tau -\overset{\wedge}{\tau}\|+\eta(\tau,\overset{\wedge}{\tau}),
$$
where $\tau,\overset{\wedge}{\tau}\in \Omega$, $\alpha$ is a constant and
$\eta(\tau,\overset{\wedge}{\tau})$ is the deviation from isometry.
The constant vector
\ben
C=\frac{\int_{\mathcal{M}}x\rhoup(x)dx}{\int_{\mathcal {M}}\rhoup(x)dx}
=\frac{\int_{\Omega}\tau\mathcal{H}(\tau)d\tau}{\int_{\Omega}\mathcal {H}(\tau)d \tau}
\enn
where $\rhoup(x)$ and $\mathcal{H}(\tau )$ are density functions of $\mathcal{M}$ and $\Omega$.
With the above notations, the following theorem is proved by Zha et al \cite{ContinuumIsomap}.

\begin{theorem}\label{thm4.1}
There is a constant vector $P_j$ such that $\phiup_j(x)=P_j^T(\tau-C)+e_j(\tau)$.
Here $e_j(\tau)=\epsilon_j^{(0)}-\epsilon_j(\tau)$ has zero mean, that is,
$\ds\int_{\Omega}\mathcal{H}(\tau)e_j(\tau)d\tau =0,$
where
\ben
\epsilon_j(\tau)&=&\frac{1}{2\lambda_j}\int_{\Omega}\eta(\tau,\overset{\wedge}{\tau})
\mathcal{H}(\overset{\wedge}{\tau})\phiup_j(x)d\overset{\wedge}{\tau},\\
\epsilon_j^{(0)}&=&\frac{1}{\int_{\Omega}\mathcal{H}(\tau)d\tau}
\int_{\Omega}\epsilon_j(\tau)\mathcal{H}(\tau)d\tau.
\enn
\end{theorem}

By Theorem \ref{thm4.1}, even if the deviation $\eta(\tau,\overset{\wedge}{\tau})$
is not zero with only a limited range of $(\tau,\overset{\wedge}{\tau})$,
then the coordinate of the low-dimensional embedding $\phiup_j(x)$ is still deformed
with the deformation being measured by $e_j(\tau)$.

In order to get a better understanding of multi-manifolds data, it
is profitable to preserve intra-manifold relationship (where $\eta(\tau,\overset{\wedge}{\tau})=0$)
and inter-manifolds relationship (where $\eta(\tau,\overset{\wedge}{\tau})\not=0$) separately.
This is because sometimes we care more about the information within the same data manifold.
Here we propose a general procedure for the design of isometric multi-manifolds learning algorithms.

Step I: The decomposition process
\begin{enumerate}
\item [1.] Cluster the whole data set. If data distribute on {\bf multiple
clusters in a manifold or manifolds, the clusters or manifolds should be identified.}
Many clustering methods can be used for this; for example, K-means, Isodata and
other methods introduced in \cite{ClusteringSurvey,ClusteringAlgorithms}.
Even if the manifolds overlay with each other, they can still be identified and
clustered \cite{ManiCluster}. At this step, the data set $X$ is clustered
into several components, and each component is considered as a data manifold.

\item [2.] Estimate parameters of data manifolds. For intrinsic dimensionality estimation,
many methods can be used: for example, the fractal based method \cite{fractal2002},
the MLE method \cite{MLE2004,revisedMLE} and the incising ball method \cite{incisingball}.
Assume that $d_m$ is the intrinsic dimension of the $m$-th data manifold.
Let $\ds d=\underset{m}{\max{d_m}}$. For the neighborhood size,
\cite{SelectionParameterIsomap} introduces a method on automatically generating parameters
for Isomap on one single data manifold. For convenience, appropriate neighborhood sizes ($k_m$ or
$\varepsilon_m$ for $X^m$) can be given manually for data manifolds.

\item [3.] Learn the data manifolds individually.
One data manifold can be learned by traditional manifold learning algorithms.
Here, we propose to rebuild a graph on each data manifold with a new neighborhood size
to better approximate the intra-manifold geodesics. In doing so, Yang's methods
\cite{kMST2004,MinkST2005,kEC2005,KVC2006} and the{\bf $k$-CG }graph construction method are preferred,
where the{\bf k-CG }graph construction method will be described later.
It is assumed that the low-dimensional embedding for $X^m$ is $Y^m$.
\end{enumerate}

Step II: The composition process
\begin{enumerate}
\item [1.] Preserve a skeleton $I$ of the whole data set in a low-dimensional space $\R^d$.
The skeleton $I$ should be carefully designed so that it can represent the global structure of $X$.
Let $RY_I$ be the low-dimensional embedding of $I$.

\item [2.] Transform $Y^m$s into a single coordinate system by referring to $RY_I$.
In order to faithfully preserve the intra-manifold relationship,
Euclidean transformations can be constructed and used.
Using the embedding points $RY^m\subset RY^I$ and the corresponding points from $Y^m$,
we can construct an Euclidean transformation from $Y^m$ to the coordinate system of $RY_I$.
\end{enumerate}

The idea of using a decomposition-composition procedure is not new, which was first
used by Wu et al. \cite{ExtendedIsomaplearning} in their split-augment process and
well developed and used in \cite{DCIsomap}. The procedure we proposed here aims to solve a more
general problem. Step I.1 permits that the designed learning algorithm has a good ability
to identify data manifolds. Step I.2 gives a guideline on learning manifolds with different intrinsic
dimensionality and neighborhood sizes. Step I.3 learns data manifolds individually
so that the intra-manifold relationship can be faithfully preserved.
{\bf Step II.1 is the most flexible part of the procedure which allows us to design
new isometric multi-manifolds learning algorithms.} A well designed skeleton $I$ can better represent the
inter-manifolds relationship. In the following subsections, we will introduce a
new multi-manifolds learning algorithm and revise the original D-C Isomap algorithm
with the help of this general procedure.

\subsection{A new algorithm for isometric multi-manifolds learning}

Based on the proposed procedure, we design a new multi-manifolds learning algorithm.
As an extension of the classical Isomap method to multi-manifolds data,
the method will be referred to as multi-manifolds Isomap or M-Isomap.
It is assumed that $X$ is also interchangeable to represent the matrix $[x_1,x_2,\cdots,x_N]$,
where $x_i,$ $i=1,\cdots,N$ are column vectors in $\R^D$.

\subsubsection{Using the k-CC method to construct a neighborhood graph and identify manifolds}

\begin{table}[t]
\caption{Computational complexity comparison of k-NN, k-MSTs, Min-k-ST, k-EC and k-VC methods,
where TC stands for time complexity and IL stands for the
time complexity for incremental learning}\label{tb1}
\begin{tabular*}{8.5cm}{llllll}
\hline
& k-NN& k-MST& Ming-k-ST & k-EC & k-VC \\
\hline
TC & $O(kN^2)$ & $O(k^2N^2)$& $O(k^2N^2)$ & $O(k^2N^2)$ &$O(N^3)$\\
IL & O(kN)     & $O(N\ln N)$ & &  & $O(N\ln N+kN)$\\
\hline
\end{tabular*}
\end{table}

Table \ref{tb1} shows the time complexity of the k-NN, K-Min-ST, k-EC and k-VC methods
on the neighborhood graph construction. As shown in the table, the k-NN method has the
lowest computational complexity $O(kN^2)$.

For incremental learning, the computational complexity of the k-NN, k-MSTs and k-VC methods
are $O(kN)$, $O(N\ln N)$ and $O(N\ln N+kN)$, respectively \cite{APSDsMST,APSDs}.
The computational complexity of the Min-k-ST and k-EC methods for incremental learning are
unavailable. For data on one single manifold, the improvement of performance of Yang's methods
becomes insignificant when the neighborhood size $k$ increases for the $k$-NN method.
More importantly, the $k$-NN method implicitly has the property of clustering to multi-manifolds data.
Data points of the same manifold tend to be connected by paths and disconnected otherwise
when each data point is connected with its neighbors by edges. Although the $k$-NN method
is not a robust clustering algorithm, it is computationally efficient for both clustering and
graph construction. Therefore, we introduce a variation of the $k$-NN method
which inherits the computational advantage of the $k$-NN method.
The method{\bf is able} to identify data manifolds and construct a
totally connected neighborhood graph. In the rest of the paper, the
proposed neighborhood graph construction method will be referred as
the $k$-edge connected graph method (the $k$-CG method).

The summary of the $k$-CG algorithm is follows. First, given a neighborhood size $k$
or $\varepsilon$, each data point is connected with its neighbors.
If the data points distribute on several clusters or manifolds,
several disconnected graphs will be constructed. Data points are assigned to the
same data manifold if there is a path connecting them on the graphs.
Then, we connect each pair of graphs by $k$ nearest pairs of data points.
For robustness of the algorithm, each data point is only allowed to
have one inter-manifolds edge at most.\\ \ \\

\begin{algorithm} ({\bf The $k$-CG algorithm:}) \\
\noindent {\bf Input:} Euclidean distance matrix $D$, whose $(i,j)$-th entry is $\|x_i-x_j\|$ and
neighborhood size $k$ or $\varepsilon.$

\noindent
{\bf Output:} Graph $G=(V,E)$, the number $M$ of clusters, label of the data.

\noindent
{\bf Initialization:} $V=\{x_1,\cdots,x_n\}$, $V'=V$, $E=\phiup$, $Queue=\phi$

\begin{algorithmic}[1]
\FOR{i=1 to $N$}
\STATE Identify nearest neighbors $\{x_{i1},\cdots,x_{il_i}\}$
for $x_i$ by $k$-nearest-neighbors or $\varepsilon$-nearest-neighbors. Let $E=E\bigcup \{e_{i1},\cdots, e_{il_i}\}$
\ENDFOR

\STATE  Set $M=1$
\WHILE{$V'$ is not empty}
\STATE $x\in V'$, in-Queue\{x\}, label(x)=$M$, $V'=V'- \{ x\}$
\WHILE{$Queue$ is not empty}
\STATE x=de-Queue
\STATE $\forall y$: y is connected with x by an edge
\IF{y is not labeled}
\STATE in-Queue\{y\}, label(y)=$M$, $V'=V'- \{ y\}$
\ENDIF
\ENDWHILE
\STATE M=M+1
\ENDWHILE
\STATE M=M-1
\IF{$M\geq$ 2}
\STATE $k=$average($\{s_i\}_{i=1}^N $ \COMMENT{ where $s_i$ is the number of neighbors of $x_i$}
\FOR{$i=1$ to $M$}
\FOR{$j=i+1$ to $M$}
\STATE Find $k$ shortest inter-manifolds edges $e_1,\cdots,e_k$ between
data manifolds $i$ and $j$ and make sure that their ending
vertices are not identical. Let $E=E\bigcup \{e_1,\cdots,e_k\}$
\ENDFOR
\ENDFOR
\ENDIF
\end{algorithmic}
\end{algorithm}

The main difference between the $k$-NN method and the $k$-CG method is lines $4$ to $25$,
which identify components (data manifolds) and connect different components of the graph.
This change makes the constructed graph totally $k$-edge connected.
Compared with the method proposed in \cite{ExtendedIsomaplearning}, the $k$-CG method
constructs a neighborhood graph with $k$ inter-manifolds edges, which is able to control the
rotation of the embedding of the data manifolds.
In Section \ref{sec5}, the $k$-CG Isomap method, which uses the $k$-CG method to construct
a totally connected graph and then perform the classical Isomap on the graph,
is compared with the M-Isomap method.
It can be easily seen that the $k$-CG Isomap suffers the limitation
which has been shown by Theorem \ref{thm4.1}.{\bf We assume that
$\{x_{n(i)}^m \}_{i=1}^k $ is the subset of $X^m$ whose data points connect with
manifold $X^{n(i)}$, $i=1,\cdots,k$.}

\subsubsection{Learn data manifolds individually}

As $X^m$ is considered as a single data manifold in $\R^D$,
it is possible to find its intrinsic parameters. The incising ball method \cite{incisingball}
is utilized to estimate the intrinsic dimensionality, which is simple to implement
and always outputs an integer result. Assume that $d$ is the highest intrinsic dimensionality
of data manifolds. The neighborhood size $k_m$ or $\varepsilon_m$ of each data manifold
is given manually and the graph on the data manifold $X^m$ is rebuilt.
It is expected that the new neighborhood graph on $X^m$ can give better approximations to the
intra-manifold geodesics. The approximated geodesic distance matrix
for $X^m$ is written as $D_m$. By applying the classical MDS on $D_m$,
the low-dimensional embedding for $X^m$ can be obtained as $Y^m=\{y_i^m\}_{i=1}^{S_m}$.

\subsubsection{Preserve a skeleton of the data manifold $X$}

First, inter-manifolds distances are computed. Assuming that $x_p^m$
and $x_q^n$ are any data points with $x_p^m\in X^m$ and $x_q^n\in X^n$,
their distance can be computed by
\be\label{e41}
d_G(x_p^m,x_q^n)=\underset{t=1\cdots k}{\min}\{d_G(x_p^m,x_{n(t)}^m)
 +\|x_{n(t)}^m,x_{m(t')}^n\|+d_G(x_{n(t')}^n,x_q^n)\},
\en
where $d_G(x_p^m,x_{n(t)}^m)$ is the shortest path on the neighborhood graph of $X^m$.
Although $d_G(x_p^m,x_{n(t)}^m)$ may not be the shortest path on the totally connected graph of $X$,
Eq. (\ref{e41}) is an efficient way to approximate distances across manifolds.
Assume that $D_{mn}$ is the distance matrix across over $X^m$ and $X^n$.
Then the furthest inter-manifolds data points are computed by
\be
\{fx_n^m,fx_m^n \}= \arg \max d_G(x_i^m,x_j^n),\qquad d_G(x_i^m,x_j^n)\in D_{mn}. \label{e42}
\en
Without loss of generality, we may assume
$$
I^m=\{x_i^m\}_{i=1}^{l_m}=\bigcup_{n=1}^M \{x_{n(1)}^m,\cdots,x_{n(k)}^m ,fx_n^m\}.
$$
Then $I=\bigcup_{m=1}^M I^m$ is considered as the global skeleton of $X$.
On the data manifold $X$, it can be seen that the skeleton $I$ formulates a sparse graph.
We assume that $D_I=(d_I(i,j))$ is the distance matrix of $I$, where
\be\label{e43}
d_I(i,j)=\left\{\begin{array}{ll}
d_G(x_i^m,x_j^n)\in D_{mn}, &  x_i \in X^m, x_j\in X^n  \\
d_G(x_i^m,x_j^m)\in D_m,  &  x_i,x_j \in X^m
          \end{array} \right.
\en\\
By applying the classical MDS algorithm on $D_I$, the low-dimensional embedding $RY_I$ of $I$
can be obtained. It is assumed that $RY_I^m=\{ry_i^m\}_{i=1}^{l_m}\subset RY_I$
is the embedding of $I^m$.

\subsubsection{Euclidean transformations}

Assume that $Y_I^m=\{y_i^m\}_{i=1}^{l_m}\subset Y^m$ and $y_i^m$ corresponds to $x_i^m$.
Then the Euclidean transformation from $Y^m_I$ to $RY_I^m$ can be constructed as follows.

The general Euclidean transformation can be written as
\ben
ry = \mathcal{A}y+\beta,
\enn
where $\mathcal{A}$ is an orthonormal matrix and $\beta$ is a translation vector.
For the $m$-th data manifold, it is assumed that the Euclidean transformation is
\ben
ry_i^m = \mathcal{A}_m y_i^m+\beta_m, \qquad\;\; i=1,\cdots,l_m.
\label{e44}
\enn
The above Euclidean transformation can be rewritten in the matrix form:
\be\label{e45}
RY_I^m=\mathcal{A}_m Y_I^m+\beta_m e^T
      =\begin{pmatrix}\mathcal{A}_m & \beta_m\end{pmatrix}
             \begin{pmatrix} Y_I^m \\e^T \end{pmatrix}
\en
where $e$ is a vector with all ones. Equation (\ref{e45}) can be solved using the least square
method, and the solution is given by
\be\label{e46}
\begin{pmatrix}\mathcal{A}_m & \beta_m \end{pmatrix}= RY_I^m
\begin{pmatrix}Y_I^m \\ e^T \end{pmatrix}^T
\left(\begin{pmatrix} Y_I^m \\e^T \end{pmatrix}
\begin{pmatrix} Y_I^m \\e^T \end{pmatrix}^T+\lambda I\right)^{-1}
\en
where $I$ is the identity matrix and $\lambda$ is a regularization parameter in the singular case.
However, the least square solution does not necessarily provide an orthonormal matrix $\mathcal{A}_m$.
We now propose to use the QR decomposition to get the orthonormal matrix $\mathcal{A}_m$.
The QR process can be written as
\be\label{e47}
\begin{pmatrix} \mathcal {A}_m & R \end{pmatrix} = QR(\mathcal{A}_m)
\en
where the diagonal elements of $R$ are forced to be nonnegative.
Then $\beta_m $ can be recomputed by minimizing the cost function
\ben\label{e48}
C(\beta_m)= \sum_{i=1}^{l_m}  \|\mathcal {A}_m y_i^m+\beta_m-ry_i^m\|^2.
\enn
Solving the equation $\ds\frac{\partial C(\beta_m)}{\partial\beta_m}=0$ gives
\be\label{e49}
\beta_m=\frac{1}{l_m}\sum_{i=1}^{l_m}(ry_i^m-\mathcal{A}_my_i^m)
\en
The low-dimensional embeddings $Y_m$ ($i=1,\cdots,M$) can be formed into a global
coordinate system using the constructed Euclidean transformations.

\subsubsection{The M-Isomap algorithm}

The detailed M-Isomap algorithm is summarized in the following table.

\begin{tabular*}{8.5cm}{ll}
\hline
{\bf Input:}  & $X=\{{x_i}\}_{i=1}^N $ with $x_i\in\mathbb{R}^D$. Initial neighborhood\\
              & size $k$ or $\varepsilon$.\\
\hline
{\bf  Step I.1} & Perform the $k$-CG algorithm on $X$. Data\\
                & manifolds $\{X^m\}_{m=1}^M$ and the set of inter-manifolds\\
                & points $\{x_{n(i)}^m\}_{i=1}^k$ of $X^m$ can be obtained. \\
{\bf Step I.2} & Estimate parameters of the data manifolds.\\
               & Assume that the intrinsic dimension $d_m$ and \\
               & neighborhood size ($k_m$ or $\varepsilon_m$) are parameters for \\
               & $X^m$. Let $\ds d=\underset{m}{max}\{d_m\}$ and rebuild the neighbor-\\
               & hood graph on $X^m$. \\
 {\bf Step I.3} & Classical Isomap algorithm is performed on $X^m$  \\
                & with new neighborhood graph,  ($m=1,\cdots,M$).\\
                & The corresponding low-dimensional embedding\\
                & of $X^m$ is denoted by $Y^m$.\\
\hline
{\bf Step II.1} & Inter-manifolds distance matrix $D_{mn}$ is computed\\
                & by Eq. (\ref{e41}); thus $\{fx_n^m\}_{m\not=n}^M$ can be found by Eq.\\
                &  (\ref{e42}). Distance matrix $D_I$ for the skeleton $I$ is\\
                & computed by Eq. (\ref{e43}). Classical MDS is performed \\
                & on $D_I$ to obtain the low-dimensional embedding  of \\
                & $I$,  which is written as $RY_I$. Assume that $RY_I^m\subset RY_I$ \\
                &   is the embedding of $I^m$.\\
{\bf Step II.2} & Construct Euclidean transformations by\\
                & Eqs. (\ref{e46})-(\ref{e49}). Use the Euclidean transformations\\
                & to transform $Y^m$ into $RY^m,$ $m=1,\cdots,M.$\\
{\bf Step II.3} & $Y=\bigcup_{m=1}^M RY^m$ is the final output.\\
\hline
\end{tabular*}

\subsection{Computational complexity of the M-Isomap algorithm}

Computational complexity is a basic issue for application. In the M-Isomap method,
the $k$-CC algorithm needs $O((k+1)N^2)$ time to construct a totally connected graph
and identify the manifolds. It needs $O(\sum_{m=1}^M S_m^2\ln S_m)$ time to compute
the shortest path on each data manifold and $O(\sum_{m=1}^M S_m^3)$ time to perform
the classical MDS on the distance matrices of data manifolds.
The time complexity of computing the shortest path across data manifolds is
$O(\sum_{m<n}^MkS_mS_n)$ and that of finding $fx^i_j,fx^j_i$ is $O(\sum_{m<n}^MS_mS_n)$.
Performing the classical MDS on the skeleton $I$ needs $O((\sum_{m=1}^M l_m)^3)$ computational time.
The time complexity of finding the least square solution and processing the QR decomposition
for $M$ data manifolds is $O(Md^3)$. Finally, transforming $Y^m$s into a single coordinate
system needs $O(d^2N)$ computational time.

Therefore, the total time complexity of the M-Isomap method is
\ben
&&O((k+1)N^2+\sum_{m=1}^M(S_m^3+S_m^2\ln S_m)+\sum_{m<n}^M(k+1)S_mS_n\\
&&\qquad +(\sum_{m=1}^M l_m)^3+ Md^3+d^2N ).
\enn
For a large data set where $N>>M$ and $N>>d$, the overall time complexity of the M-Isomap
algorithm can be approximated by
\ben
O((k+1)N^2+\sum_{m=1}^M (S_m^3+S_m^2\ln S_m ) +\sum_{m<n}^M(k+1)S_mS_n).
\enn

\subsection{The revised D-C Isomap method}

\begin{figure}
\begin{center}
\setlength{\unitlength}{1mm}
\begin{picture}(80,40)
\put(10,0){(a)}

\put(60,0){(b)}

\qbezier(3,3)(20,3)(20,5) \qbezier(3,30)(20,30)(20,28)
\put(3,3){\line(0,1){27}} \put(20,5){\line(0,1){23}}
\put(20,25){\circle*{1}} \put(20,5){\circle*{1}}
\put(25,35){\circle*{1}} \put(23,0){\circle*{1}}
\put(20,25){\shortstack[r]{$nx_1^1$}}
\put(20,5){\shortstack[r]{$nx_2^1$}}
\put(25,35){\shortstack[r]{$nx_1^2$}}
\put(23,0){\shortstack[r]{$nx_1^3$}}
\put(20,25){\line(1,2){5}} \put(20,5){\line(3,-5){3}}
\qbezier[20](15,16)(17.5,20.5)(20,25)
\qbezier[20](15,16)(17.5,10.5)(20,5)
\qbezier[30](25,35)(24,17.5)(23,0)
\put(15,16){\circle{1}} \put(11,16){\shortstack[l]{$O_1$}}
\qbezier(45,27)(60,27)(60,30) \qbezier(45,5)(60,5)(60,3)
\put(45,5){\line(0,1){22}} \put(60,3){\line(0,1){27}}
\put(60,8){\circle*{1}} \put(60,25){\circle*{1}}
\put(72,12){\circle*{1}} \put(75,20){\circle*{1}}
\put(60,8){\shortstack[r]{$nx_3^1$}}
\put(60,25){\shortstack[r]{$nx_2^1$}}
\put(72,12){\shortstack[r]{$nx_1^3$}}
\put(75,20){\shortstack[r]{$nx_1^2$}}
\put(60,8){\line(3,1){12.6}} \put(60,25){\line(3,-1){15.8}}
\qbezier[20](45,17)(50.5,21)(60,25)
\qbezier[20](45,17)(50.5,12.5)(60,8)
\qbezier[15](75,20)(74,16)(73,12)
\put(45,17){\circle{1}} \put(40,17){\shortstack[l]{$O_1$}}
\end{picture}
\caption{Two basic cases of the relationship between the center of each cluster or manifold
and the inter-manifolds points}\label{fig1}
\end{center}
\label{fig10}
\end{figure}

D-C Isomap applies the decomposition-composition procedure.
Therefore, it is able to preserve intra-cluster distances correctly.
However, this method suffers from several limitations.
In the following, the original D-C Isomap algorithm will be revised
to overcome its limitations.

\subsubsection{Selection of centers}

D-C Isomap implicitly assumes that the inter-cluster point $nx_n^m$
is on the line which connects centers $O_m$ and $nx_m^n$. Thus it is
more sensible that $O_m$ is chosen by referring to the inter-cluster
points. Fig. \ref{fig1} illustrates two basic cases about the relationship of
the center and inter-cluster points. Although the points $nx_1^1$,
$nx_1^2$, $nx_1^3$, $nx_2^1$ and $O_1$ do not have to really lie on
the same plane in the ambient space. It is assumed that these points
formulate a triangle in the low-dimensional space. Fig. \ref{fig1} (a) shows
the case when $\angle nx_1^1nx_1^2nx_1^3+\angle nx_2^1 nx_1^3nx_1^2<180\textordmasculine$.

In the triangle $\Delta O_1nx_1^3nx_1^2$, the edge $d(nx_1^2,nx_1^3)$
can be computed as $\|nx_1^2-nx_1^3\|$. We also have
\ben
\angle O_1nx_1^2nx_1^3&=&\arccos\frac{<nx_1^1-nx_1^2,nx_1^3-nx_1^2>}{\|nx_1^1-nx_1^2\|\|nx_1^3-nx_1^2\|}\\
\angle O_1nx_1^3nx_1^2&=&\arccos\frac{<nx_2^1 -nx_1^3,nx_1^2-nx_1^3>}{\|nx_2^1-nx_1^3\|\|nx_1^2-nx_1^3\|}
\enn
Subsequently, the length of edges $d(O_1,nx_1^2)$ and $d(O_1,nx_1^3)$ can be
calculated by the Law of Sines in the triangle $\Delta O_1nx_1^3nx_1^2$.
Suggested distances between the center $O_1$ to the inter-cluster points can be calculated as
\ben\label{e51}
d'(O_1,nx_1^1)&=& d(O_1,nx_1^2)-\|nx_1^2-nx_1^1\|\\
d'(O_1,nx_2^1)&=& d(O_1,nx_1^3)-\|nx_1^3-nx_1^1\|.
\enn
For a cluster with the intrinsic dimension $2$, it is sufficient to
estimate the position of $O_1$ in the cluster by solving the following optimization problem:
\be\label{e52}
O_1 = \underset{o\in X_1}{\arg\min} f(o),
\en
where
\ben
f(o)=\sum_{i=1}^2 \|d(O_1,nx_i^1)-d'(O_1,nx_i^1)\|.
\enn
Here, $d(O_1,nx_i^1)$ is the length of the shortest path between $O_1$ and $nx_1^1$
on the graph $X^1$. For a cluster with intrinsic dimension $d_m$, at least $d_m$ distances
$d'(O_1,nx_i^1),$ $i=1,\cdots,d_m,$ are needed to estimate the position of the center $O_1$.
In this case, $f(o)$ is given by
\ben\label{e53}
f(o)=\sum_{i=1}^{d_m} \|d(O_1,nx_i^1)-d'(O_1,nx_i^1)\|
\enn

If we can not find sufficiently many distances $d'(O_1,nx_i^1)$ to locate the center,
there must be many inter-cluster points located in space, as illustrated in Fig. \ref{fig1} (b).
In this case, and when the center $O_1$ is never on the line passing through $nx_2^1nx_1^2$
and $nx_3^1nx_1^3$, we have
$$
\angle nx_1^1nx_1^2nx_1^3+\angle nx_2^1nx_1^3nx_1^2\geq 180\textordmasculine.
$$
In order to get a better preservation of the inter-cluster relationship, $O_1$ should be placed
as far as possible from these inter-cluster points. For a cluster with intrinsic dimension $2$,
it is suggested that $O_1$ should be chosen as
\be\label{e54}
O_1=\arg\underset{o\in X_1}{\max}\{g(o)\}
\en
where
\ben\label{e55}
g(o)= d(o,nx_1^1)+d(o,nx_2^1)-\|d(o,nx_1^1)-d(o,nx_2^1)\|
\enn
If the intrinsic dimension of $X^1$ is $d_m$ and $\{nx_i^1,i=1,\cdots,d_m\}$
is the set of inter-cluster points in $X^1$, then the function $g(o)$ should be given as
\ben\label{e56}
g(o)= \sum_{i< j}^{d_m}\left(d(o,nx_i^1)+d(o,nx_j^1)-\| d(o,nx_i^1)-d(o,nx_j^1)\|\right)
\enn

\subsubsection{Degenerate and unworkable cases}

\begin{figure}
\begin{center}
\setlength{\unitlength}{1mm}
\begin{picture}(80,40)
\qbezier(12,34)(10,3)(10,5) \qbezier(10,5)(27,6)(30,7)
\qbezier(30,7)(32,32)(31,36) \qbezier(31,36)(10,30)(12,34)
\qbezier(50,3)(62,8)(65,7) \qbezier(65,7)(60,30)(63,34)
\qbezier(63,34)(32,36)(49,37) \qbezier(49,37)(35,2)(50,3)
\put(31,26){\circle*{1}} \put(30,12){\circle*{1}}
\put(45,25){\circle*{1}} \put(42,10){\circle*{1}}
\put(31,26){\shortstack[r]{$nx_1^1$}}
\put(30,12){\shortstack[r]{$nx_2^1$}}
\put(45,25){\shortstack[r]{$nx_1^2$}}
\put(42,10){\shortstack[r]{$nx_2^2$}}
\qbezier(31,26)(38,25.5)(45,25) \qbezier(30,12)(36,11)(42,10)
\put(38,25.5){\circle*{1}} \put(36,11){\circle*{1}}
\put(38,25.5){\shortstack[r]{$m_1$}}
\put(36,11){\shortstack[r]{$m_2$}}
\qbezier[50](36,11)(38,23)(40,35)
\put(15,20){\shortstack[r]{$X^1$}}
\put(55,20){\shortstack[r]{$X^2$}}
\put(38,36){\shortstack[r]{$X^3$}} \put(40,35){\circle*{1}}
\end{picture}
\caption{An illustration of how to add a new cluster for D-C Isomap
algorithm.}\label{fig2}
\end{center}
\label{fig11}
\end{figure}

Since the original D-C Isomap algorithm relies on the position of the center of each cluster
to preserve the inter-cluster relationship, the algorithm does not work under certain circumstances.
Consider a simple case of two data clusters with the intrinsic dimension $d=2$.
The method does not work in this case because it implicitly requires an another data cluster
to provide sufficient rotation reference data points. Since the low-dimensional embedding
of each cluster is relocated by referring to the position of the center of each cluster.
In the case when there are three or more clusters and their centers are nearly on a line,
the original D-C Isomap can not find the exact rotation matrix.

This issue is solved in this paper by adding fictitious clusters.
The algorithm applies a trial and error procedure to determine the position
of the fictitious clusters. As an example, we now consider the case of two clusters.
As shown in Fig. \ref{fig2}, the nearest couple of inter-cluster points of the clusters $X^1$ and
$X^2$ are assumed to be $nx_1^1$ and $nx_1^2$, and $m_1$ is the middle point between $nx_1^1$ and $nx_1^2$.
The second nearest couple of inter-cluster points are $nx_2^1$ and $nx_2^2$, and $m_2$ is the middle
point between them. The third fictitious cluster $X^3$ is then suggested to be given by
\ben\label{e61}
X^3=m_1+\gamma\|nx_1^1-nx_1^2\|\frac{m_2-m_1}{\|m_2-m_1\|},
\enn
where the parameter $\gamma$ can be decided by a trial and error procedure.
Given a positive value $\beta>1$, $X^3$ is assumed to satisfy that
\be\label{e62}
\frac{1}{\beta }<\frac{\|X^3-X^1\|}{\|X^3-X^2\|}<\beta,
\en
where $\|X^3-X^1\|$ is the shortest distance between the data points from clusters $X^1$ and $X^3$.
If condition (\ref{e62}) is not satisfied, then $\gamma$ can be chosen in a pre-given range such as
$$
\{\cdots,-3,-2,-1,-\frac{1}{2},-\frac{1}{3},\cdots,\frac{1}{3},\frac{1}{2},1,2,\cdots\}.
$$

In the case when there are $M$ clusters in the data set with $M<d+1$,
we can start from the couple of clusters with the maximum nearest inter-cluster distance.
Assume that $X^1$ and $X^2$ satisfy that
\ben\label{e63}
\|X^1-X^2\|=\underset{i}{\max}\underset{j}{\min}\|X^i-X^j\|
\enn\label{e64}
with $nx_1^1$, $nx_2^1$, $m_1$, $nx_1^2$, $nx_2^2$, $m_2$ being defined as above.
Then the $(M+1)$-th cluster $X^{M+1}$ can be generated as
\ben
X^{M+1}=m_1 +\gamma\|nx_1^1-nx_1^2\|\frac{m_2-m_1}{\|m_2-m_1\|}
\enn
If $X^p$ and $X^q$ are the two nearest clusters of $X^{M+1}$, then, given $\beta >0$,
it is assumed that $X^{M+1}$ should satisfy
\ben\label{e65}
\frac{1}{\beta }< \frac{\|X^{M+1}-X^p\|}{\|X^{M+1}-X^q\|}<\beta.
\enn
If $M+1<d+1$, then replace $M$ by $M+1$ and repeat the generating procedure presented above.

PCA can be used to find the dimensionality of the subspace on which the centers are lying.
If the dimensionality of the subspace is smaller than $d$, then fictitious clusters
should be added until the centers of the clusters can anchor a $d$-dimensional simplex.

\subsubsection{The revised D-C Isomap algorithm}

The revised D-C Isomap algorithm can be given as follows.

\begin{tabular*}{8.5cm}{ll}
\hline
{\bf Input:} & $X=\{{x_i}\}_{i=1}^N,$ with $x_i\in\mathbb{R}^D$.
               Initial neighborhood\\
             & size $k$ or $\varepsilon.$\\
\hline
{\bf Step I.1} & Same as Step I.1 of the original D-C \\
               & Isomap  algorithm. \\
{\bf Step I.2} & Estimate the parameters, intrinsic dimension\\
               & $\{d_m\}_{m=1}^M $ and neighborhood sizes ($\{k_m\}_{m=1}^M$ or\\
               & $\{\varepsilon_m\}_{m=1}^M$), of the clusters.
                 Let $d=\underset{m}{\max}\,d_m$ and\\
               & rebuild the neighborhood graph for each\\
               & cluster.\\
{\bf Step I.3-4} & Same as Step I.2-3 of the original D-C\\
                 &  Isomap algorithm.\\
\hline
{\bf Step II.1} & Centers of the clusters are computed by (\ref{e52}) or\\
                & (\ref{e54}). Fictitious clusters should be added until\\
                & centers of the clusters can anchor a\\
                & $d$-dimensional simplex.\\
{\bf Step II.2-4} & Same as Step II.2-4 of the original D-C\\
                  & Isomap. Assume that $Y^m$ is transformed\\
                  & into $TY^m$.\\
{\bf Step II.5} & $Y=\bigcup_{m=1}^M TY^m$ is the final output.
\\\hline
\end{tabular*}

\section{Experiments}\label{sec5}

\subsection{3-D data sets}

\begin{figure*}
\includegraphics[width=5.5cm]{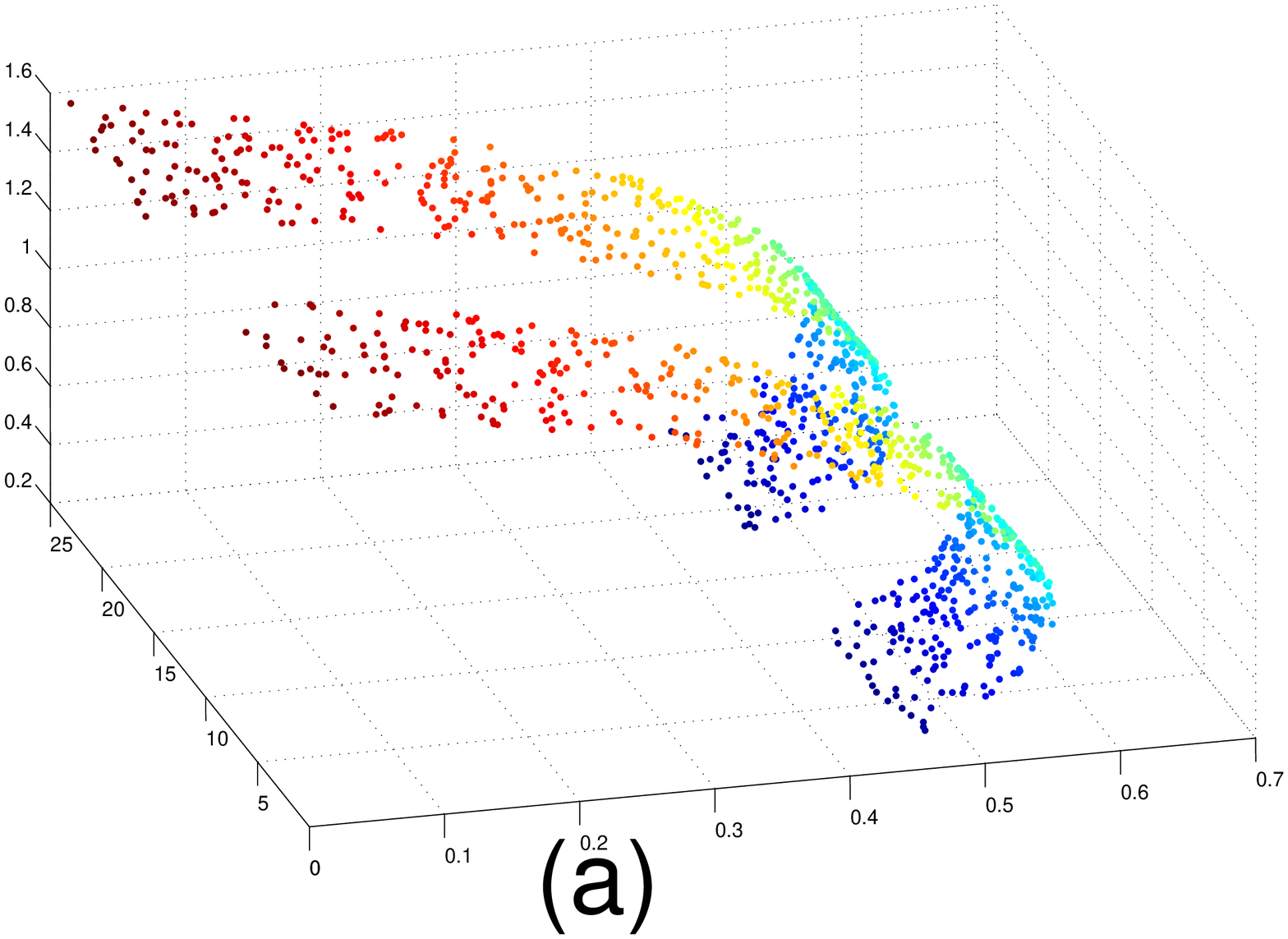}
\includegraphics[width=5.5cm]{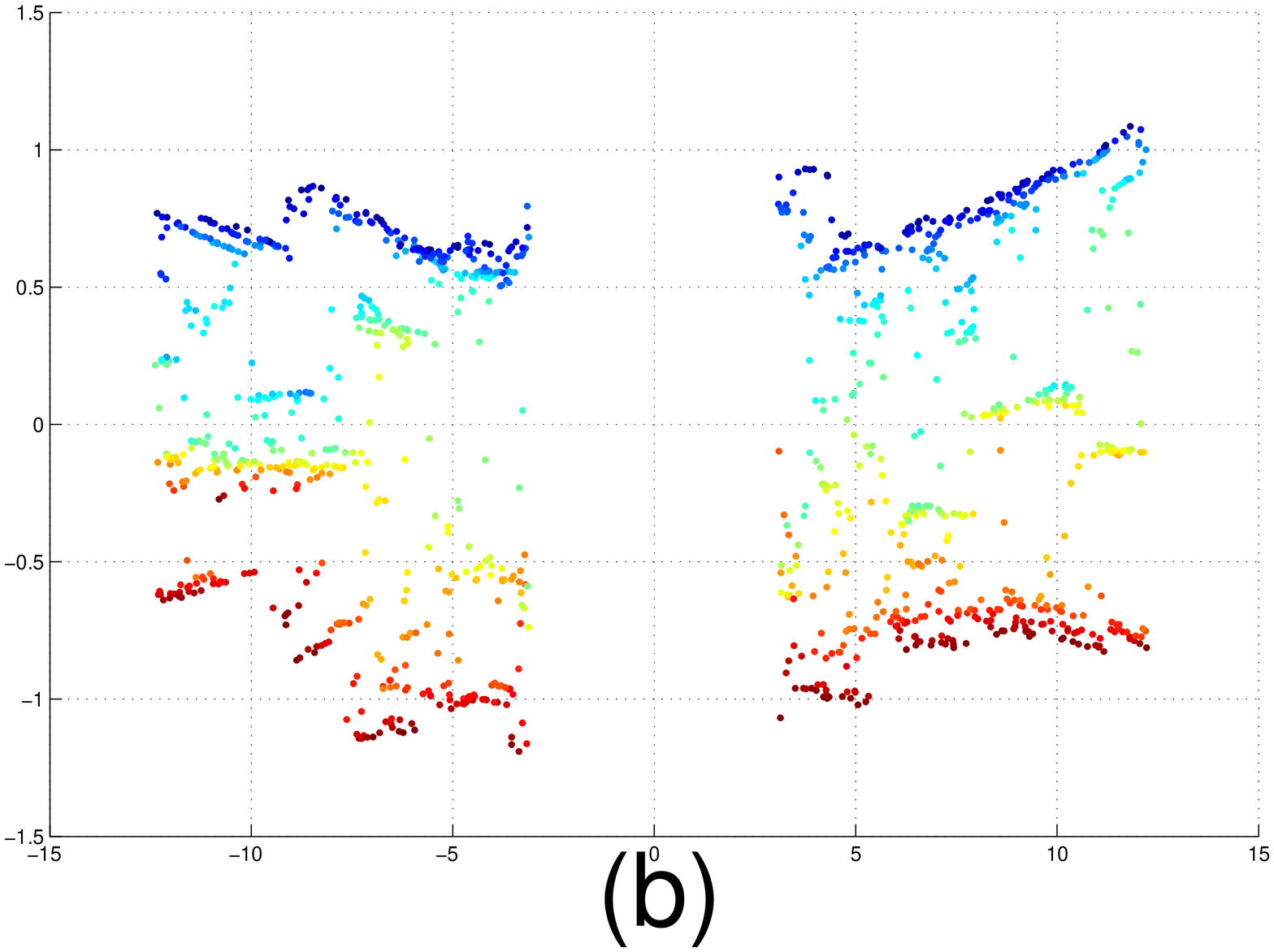}
\includegraphics[width=5.5cm]{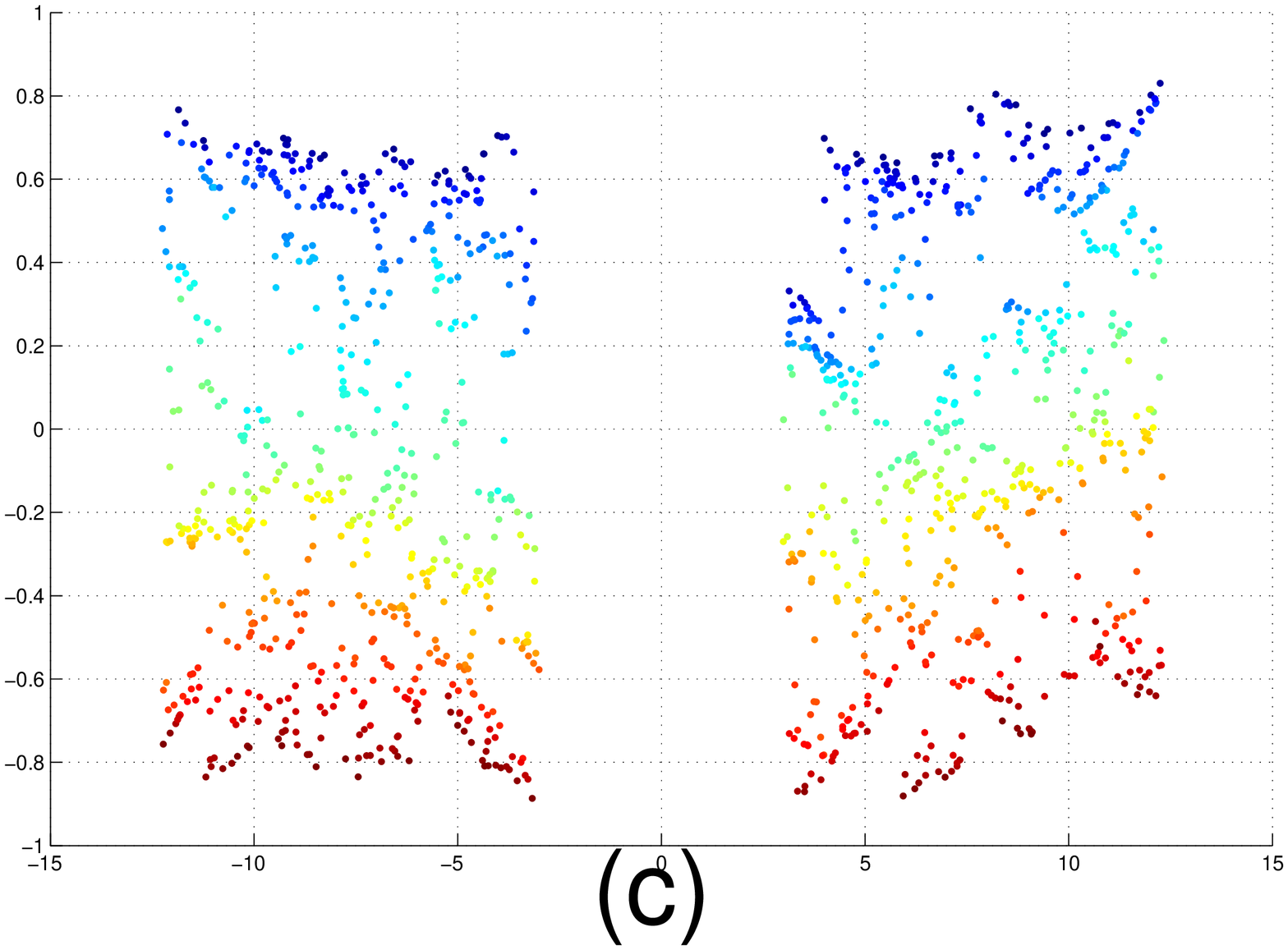}
\includegraphics[width=5.5cm]{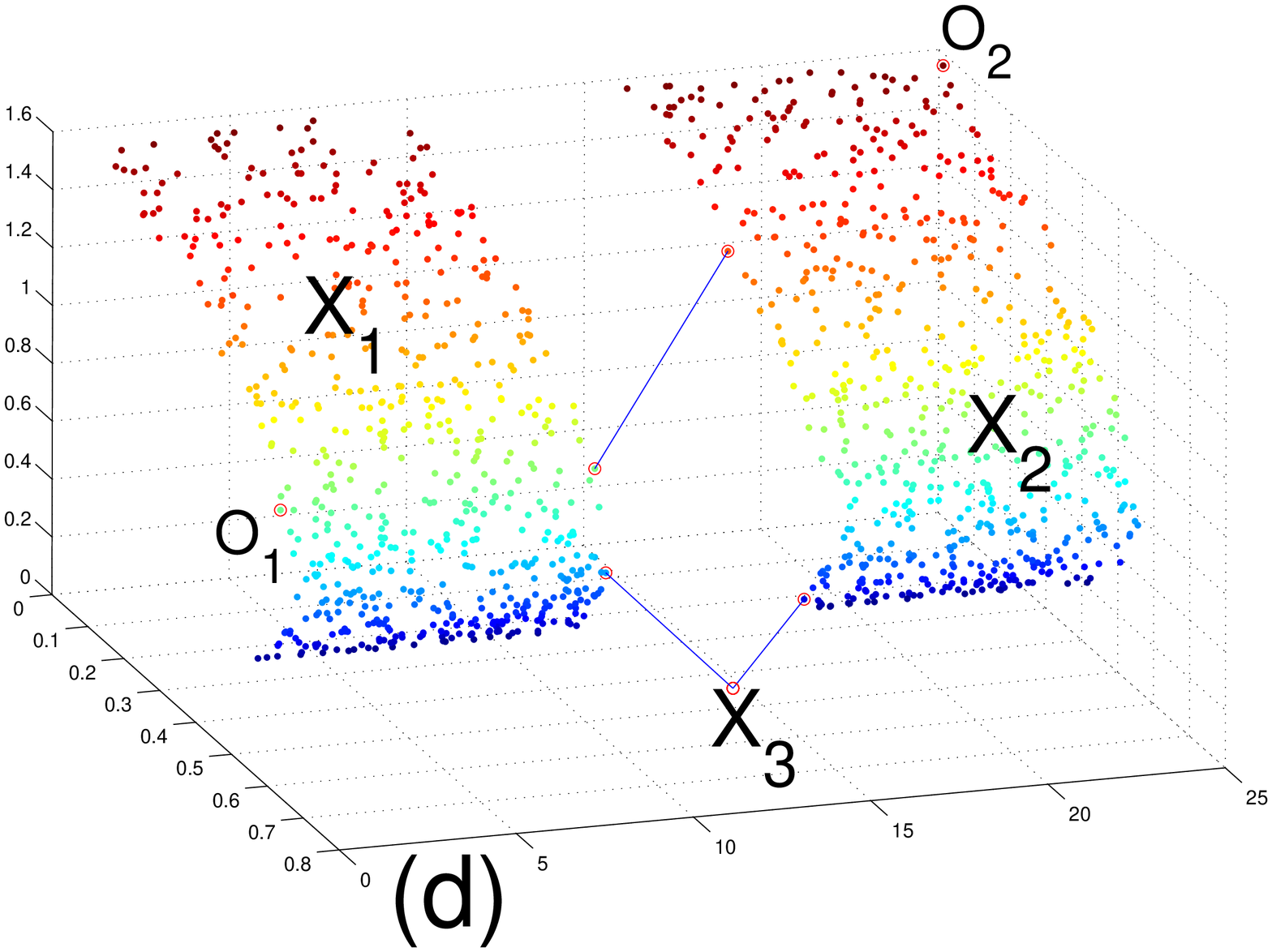}
\includegraphics[width=5.5cm]{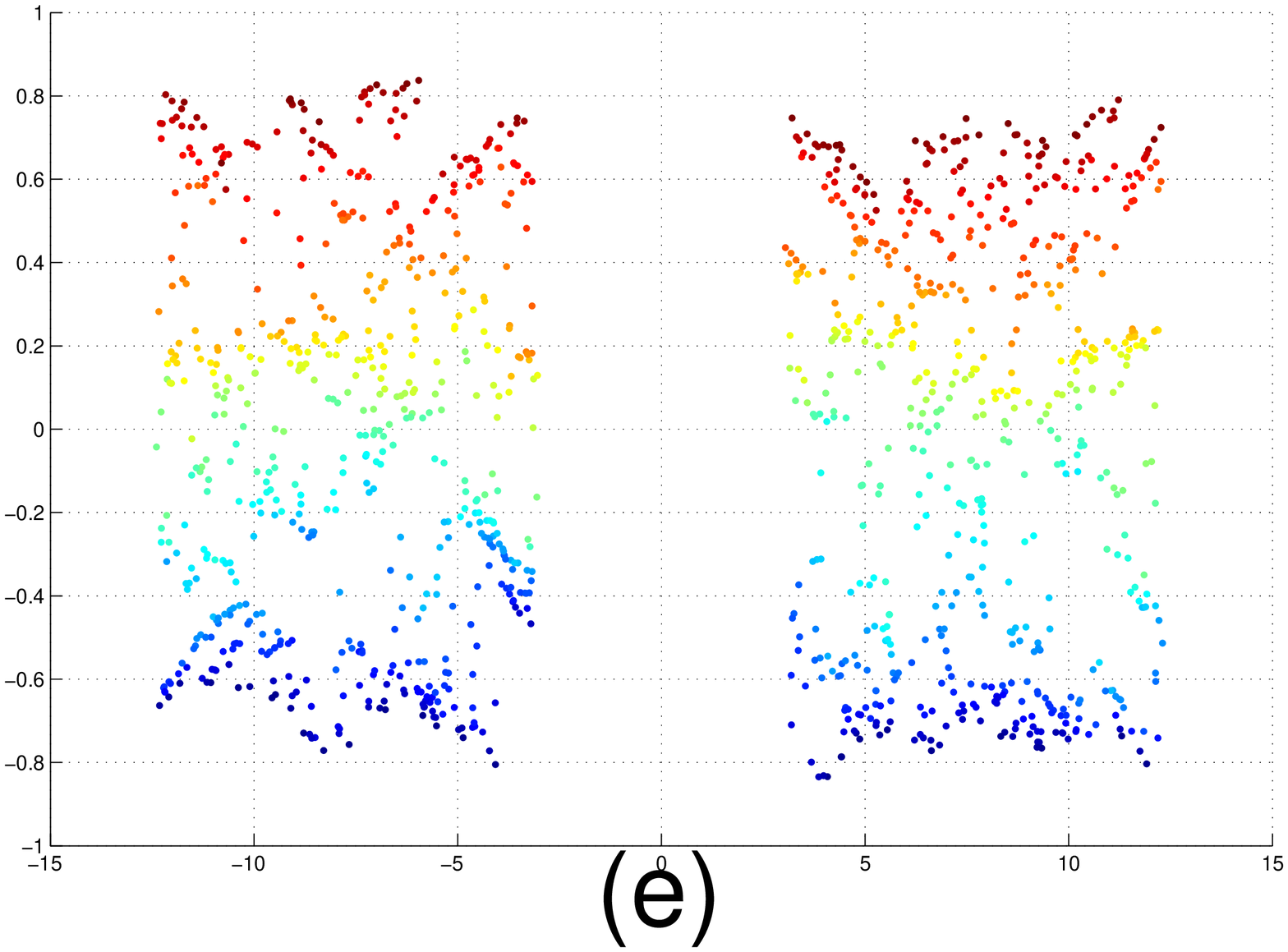}
\caption{Experiments on a two-manifolds data set. (a) The original
data set. (b) The result obtained by the k-CC Isomap. (c) The result obtained by the M-Isomap.
(d) Illustration of the procedure of the revised D-C Isomap. (e) The result obtained by
the revised D-C Isomap.
}\label{fig3}
\end{figure*}

\begin{figure*}
\includegraphics[width=5.5cm]{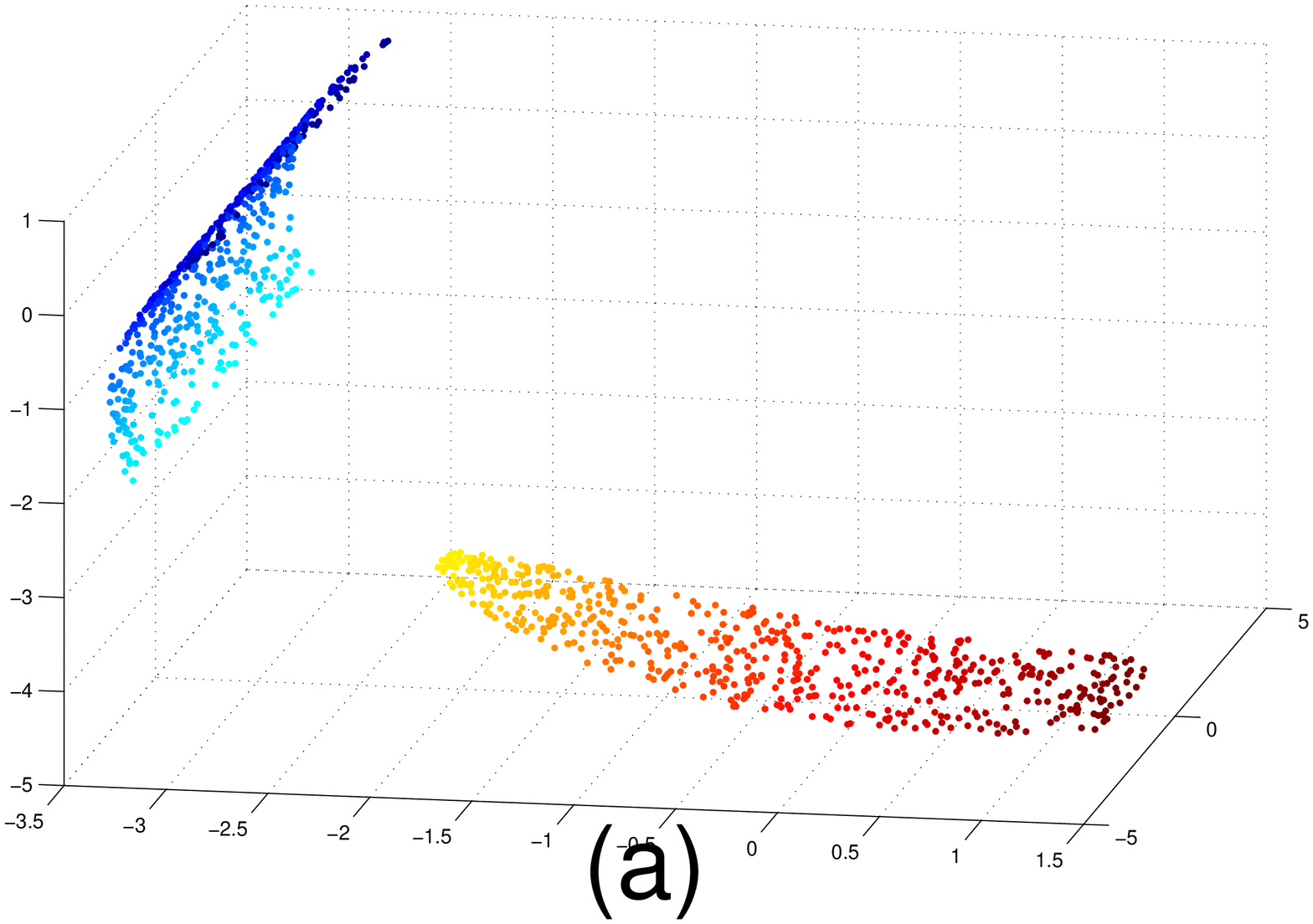}
\includegraphics[width=5.5cm]{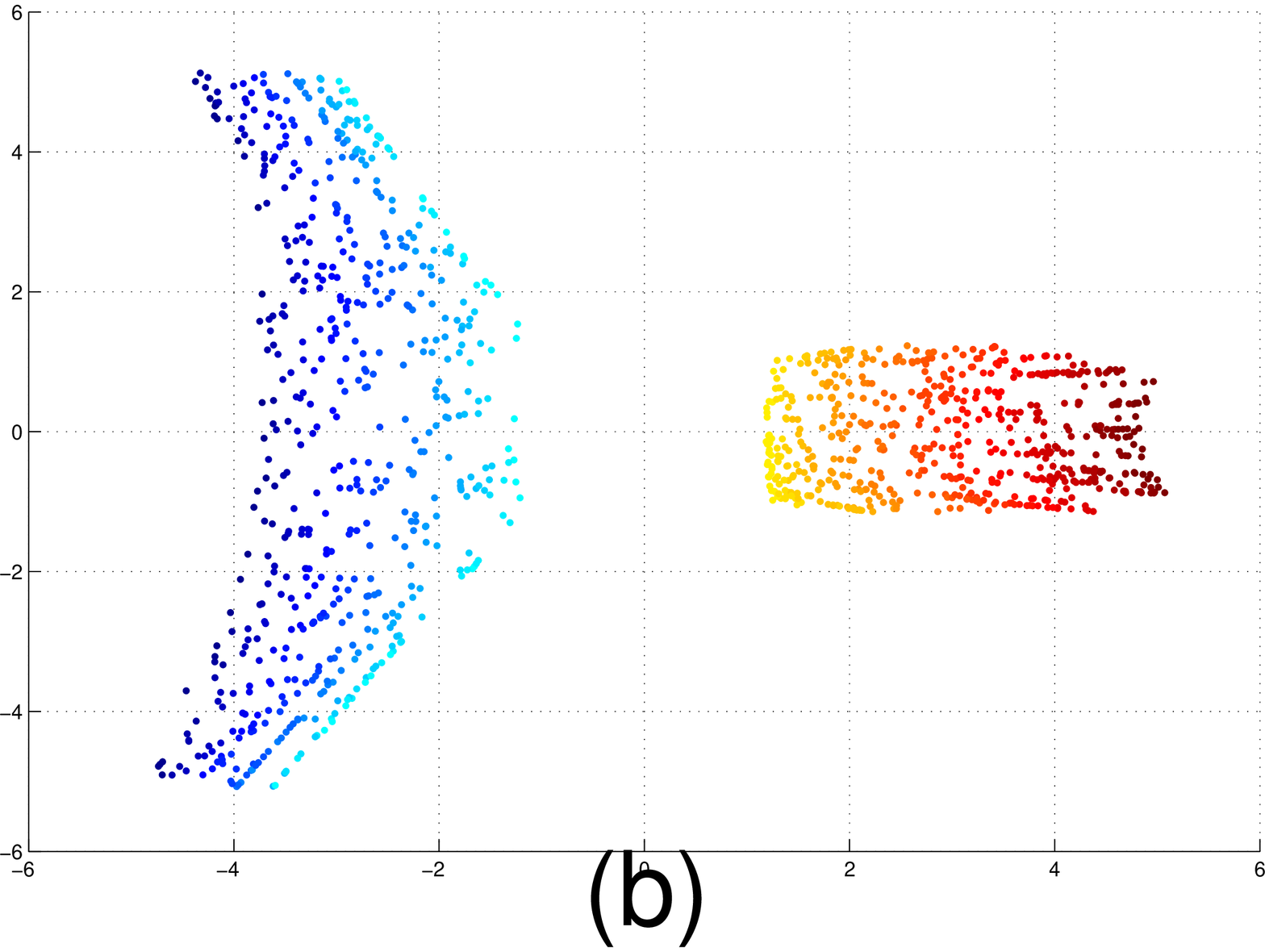}
\includegraphics[width=5.5cm]{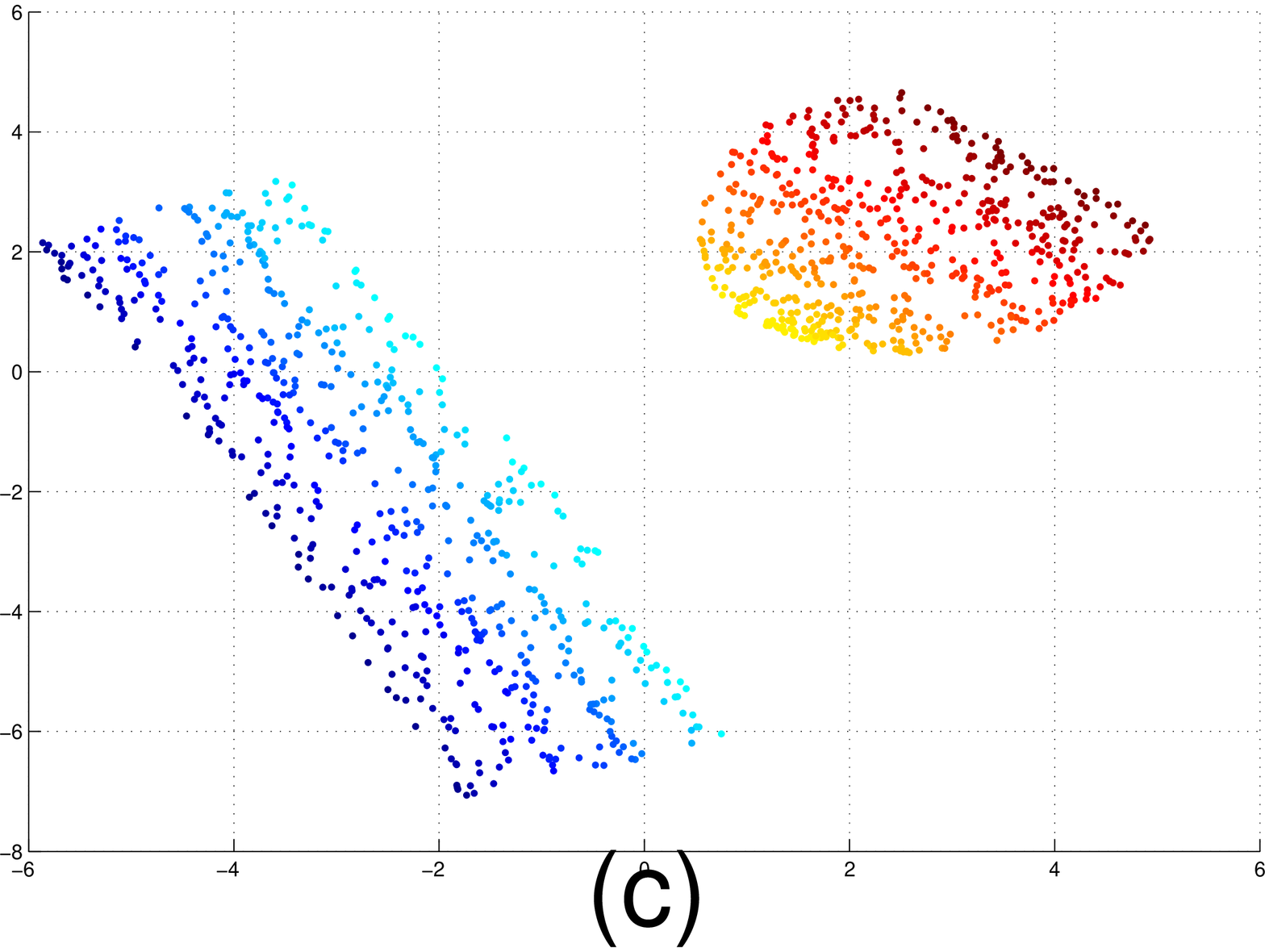}
\includegraphics[width=5.5cm]{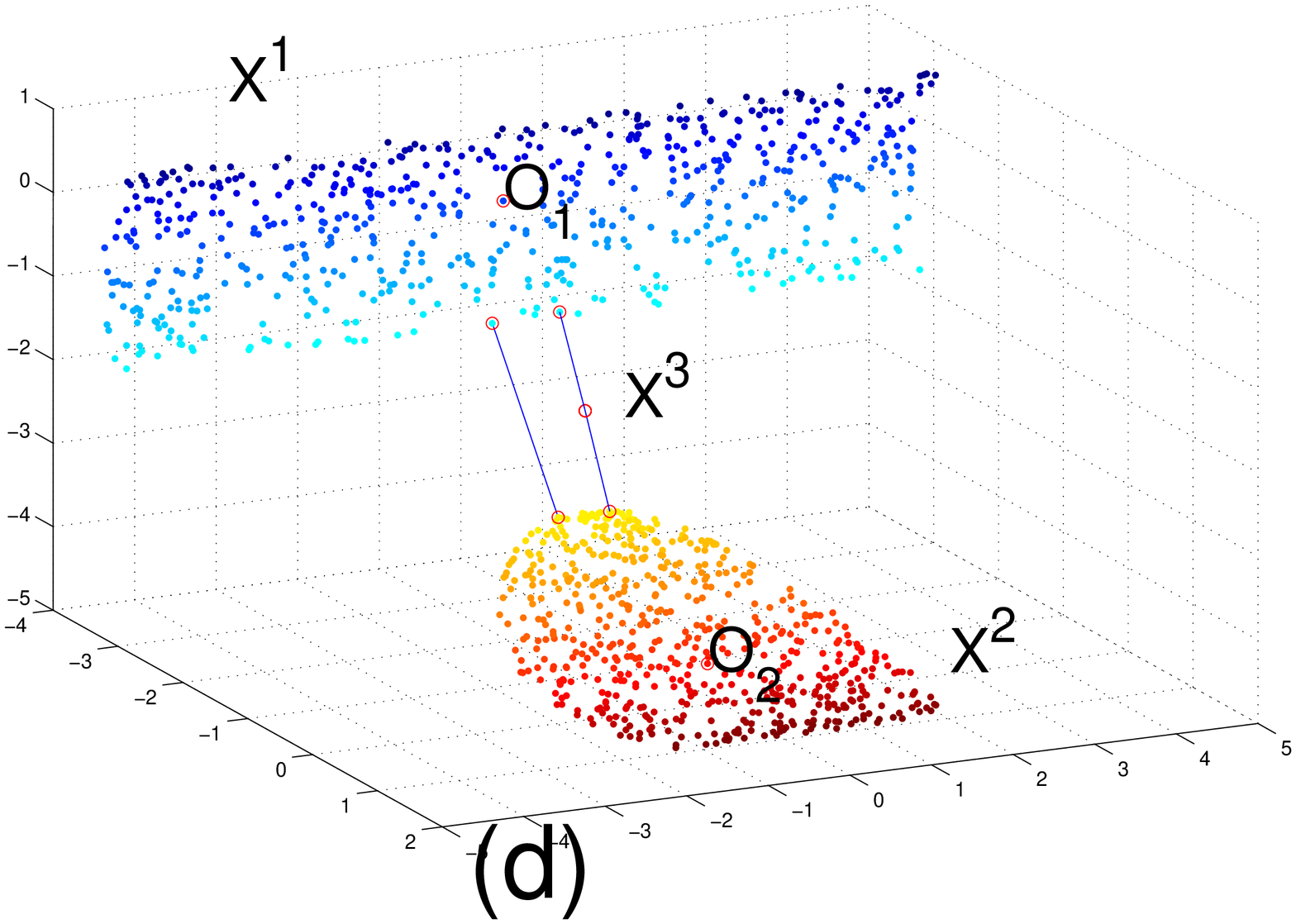}
\includegraphics[width=5.5cm]{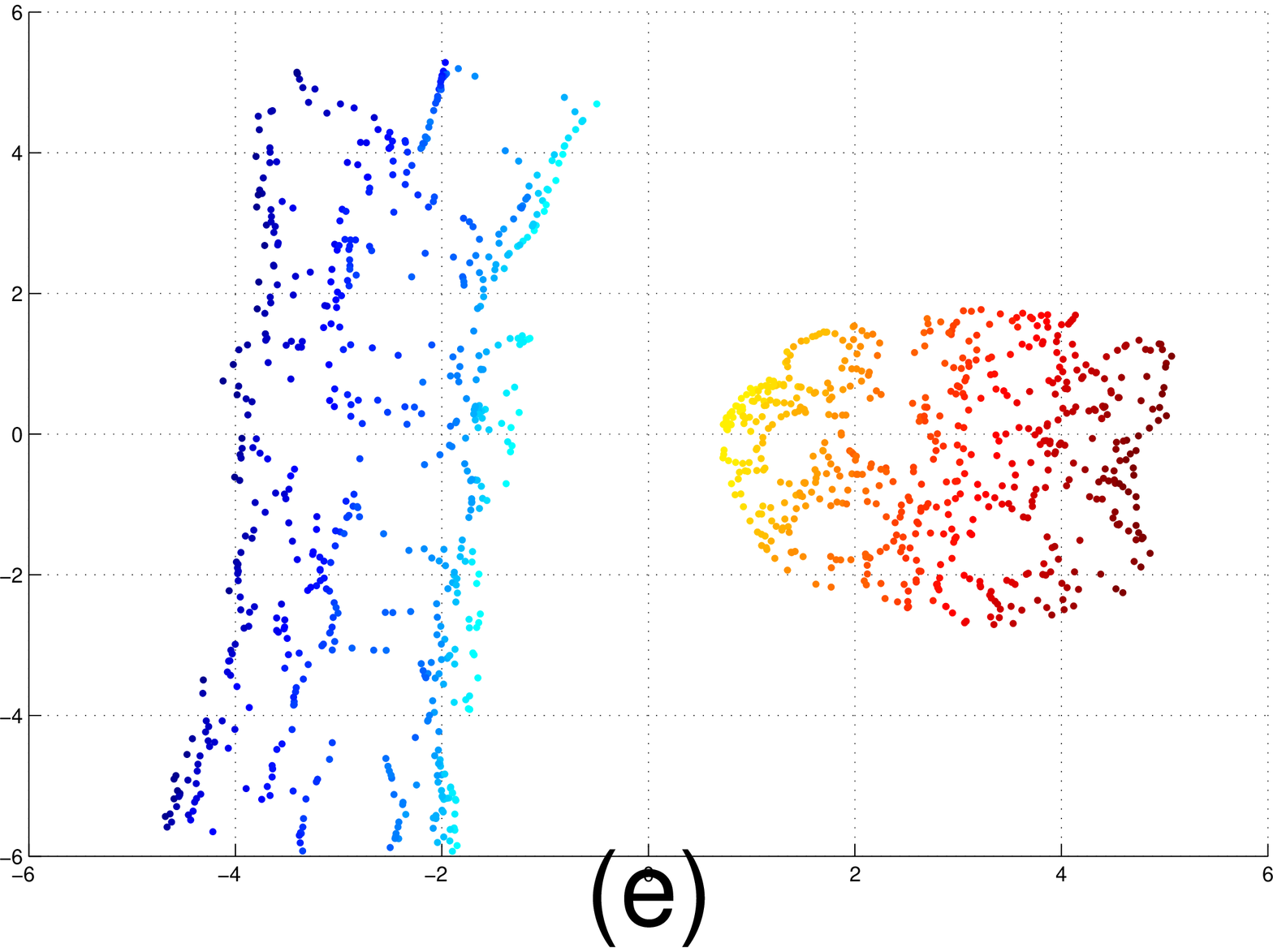}
\caption{Experiments on a two-manifolds data set. (a) The original
data set. (b) The result obtained by the k-CC Isomap. (c) The result obtained by the M-Isomap.
(d) Illustration of the procedure of the revised D-C Isomap.
(e) The result obtained by the revised D-C Isomap.
}\label{fig4}
\end{figure*}

\begin{figure*}
\includegraphics[width=5.5cm]{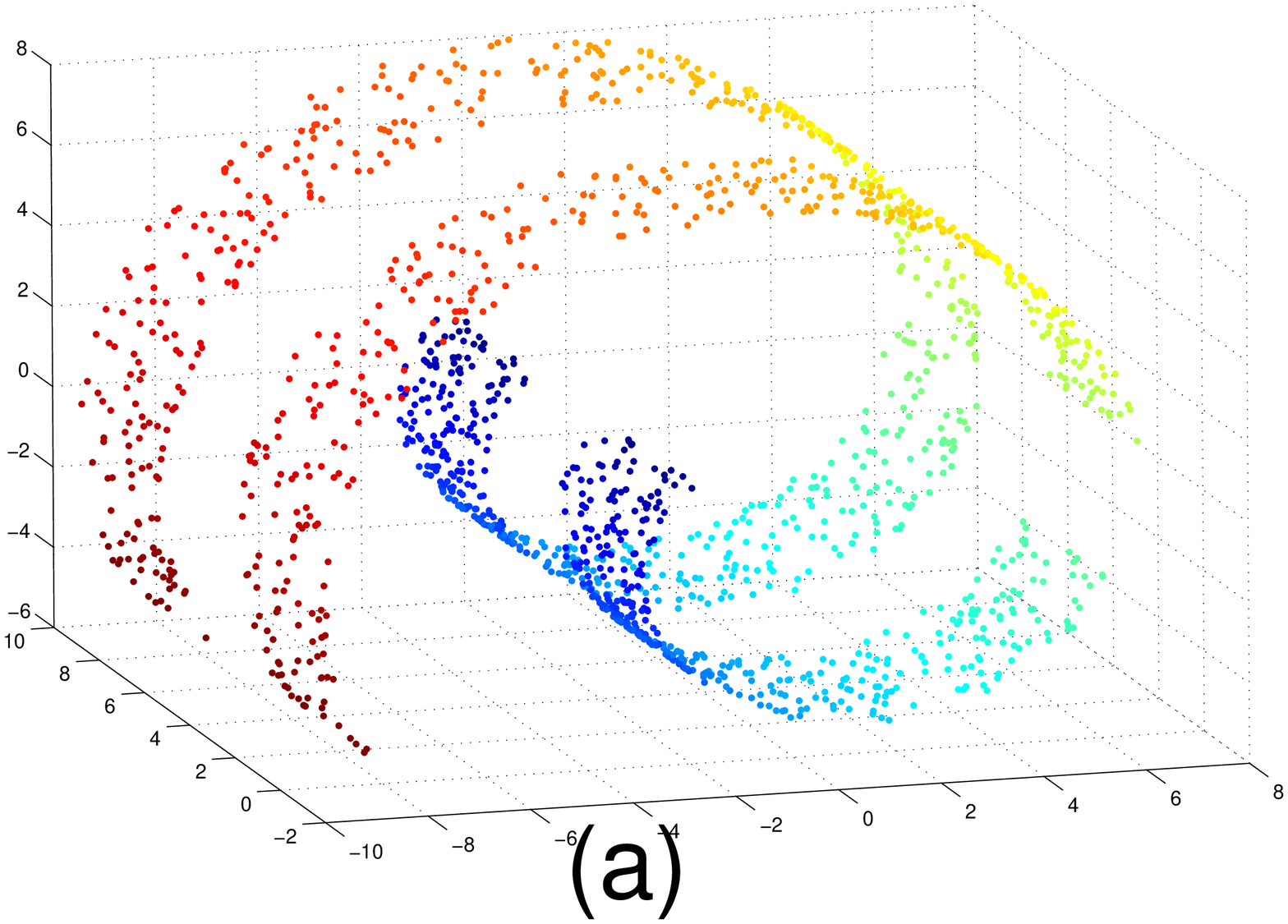}
\includegraphics[width=5.5cm]{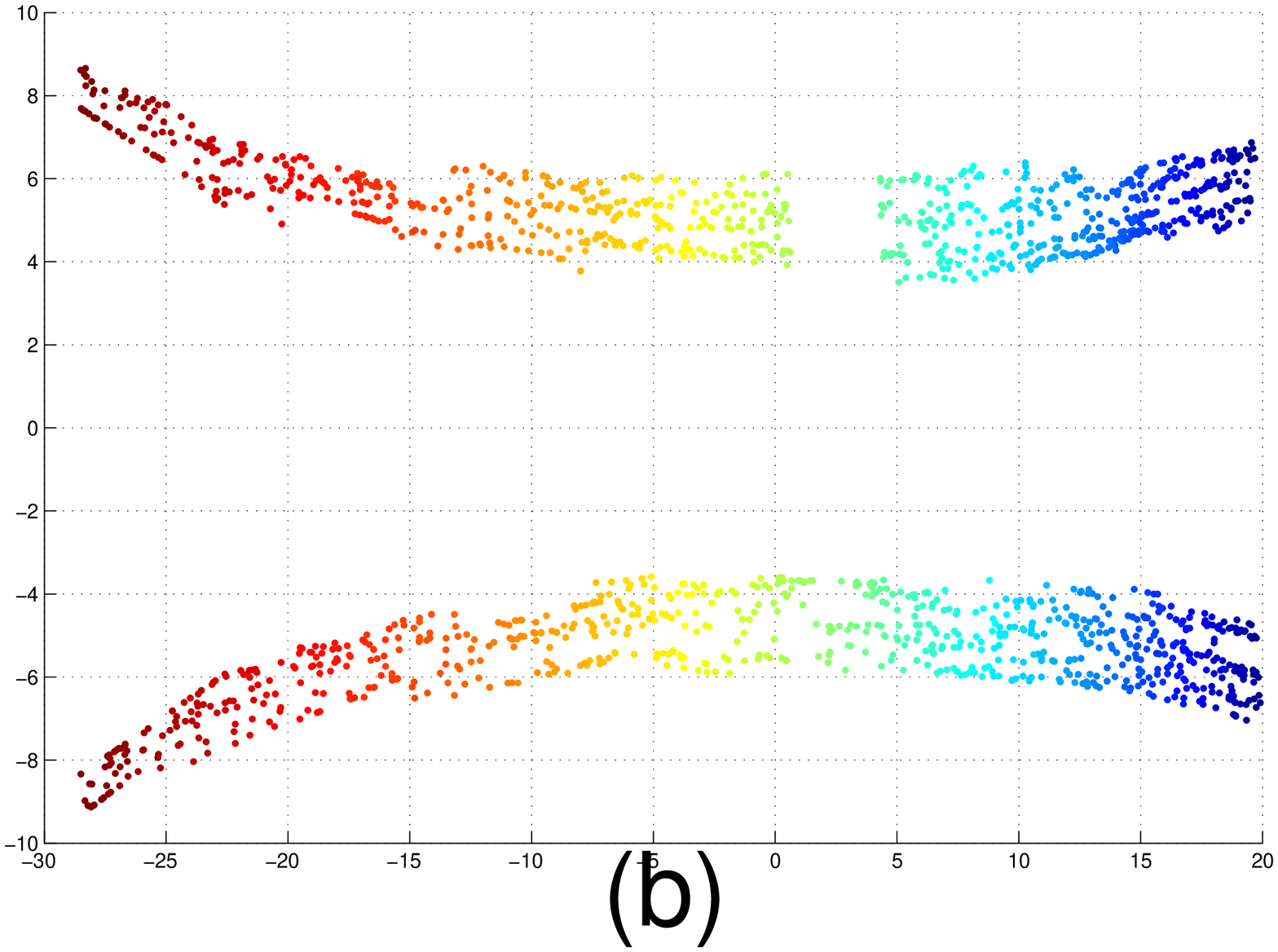}
\includegraphics[width=5.5cm]{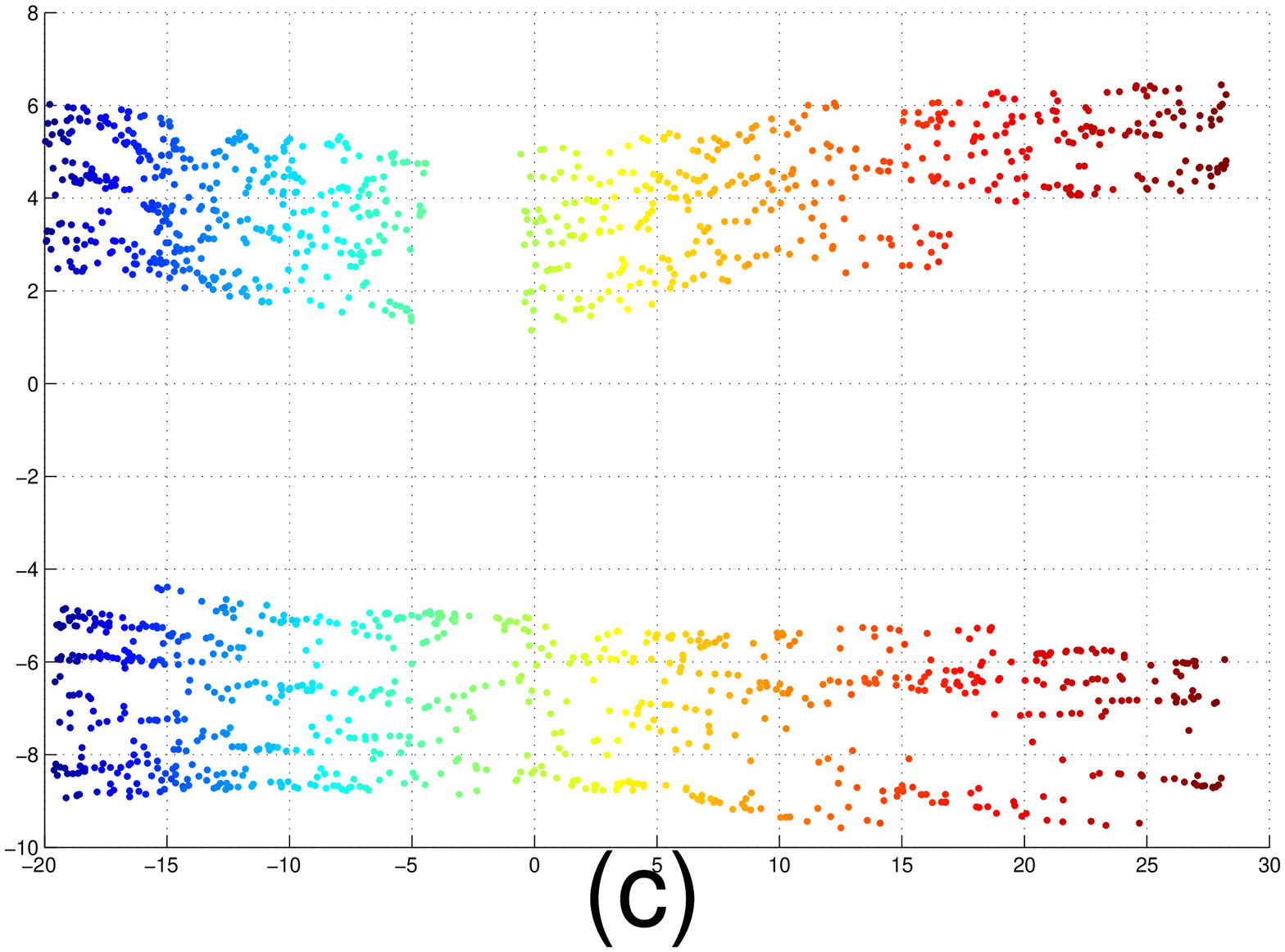}
\includegraphics[width=5.5cm]{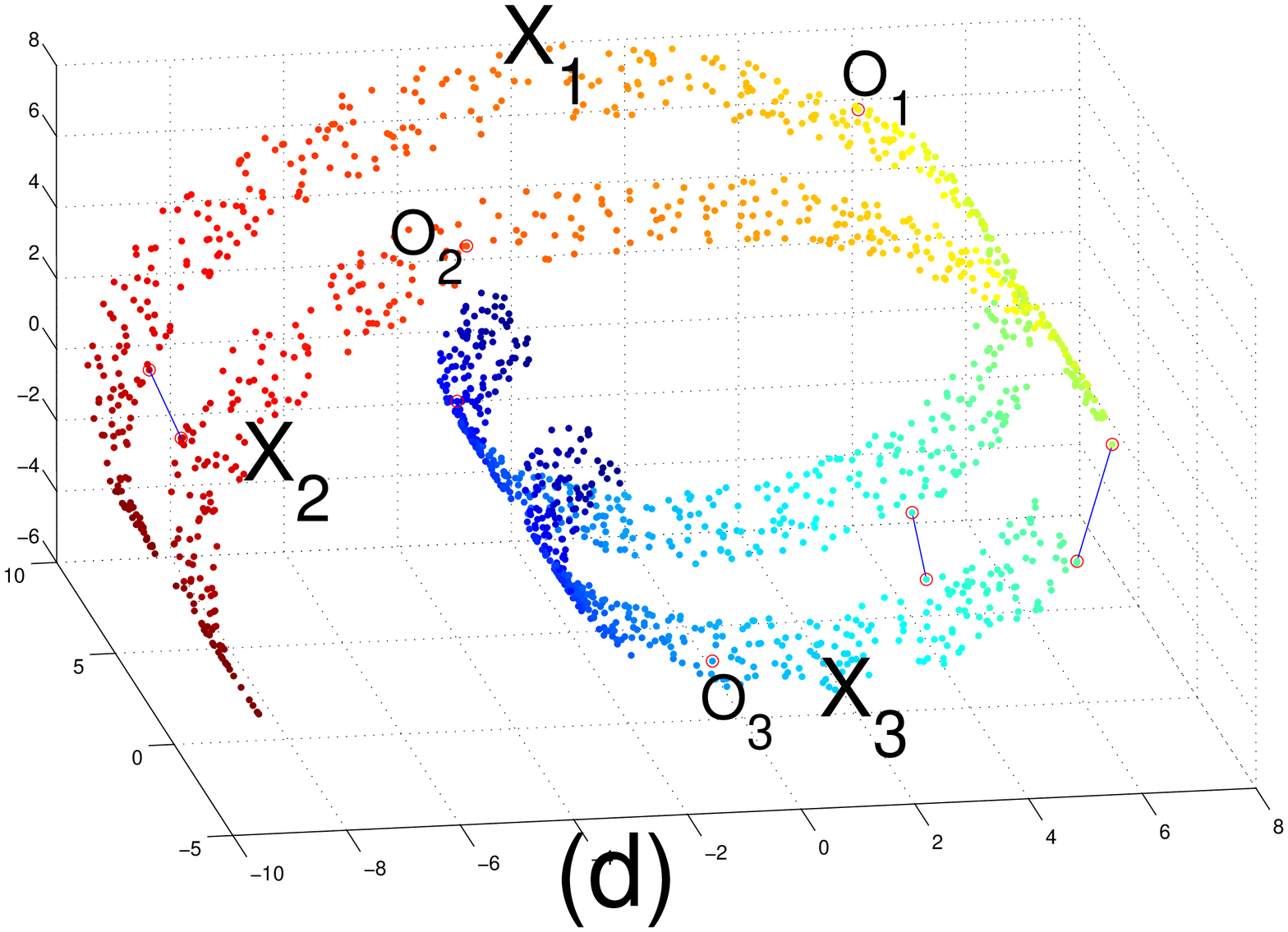}
\includegraphics[width=5.5cm]{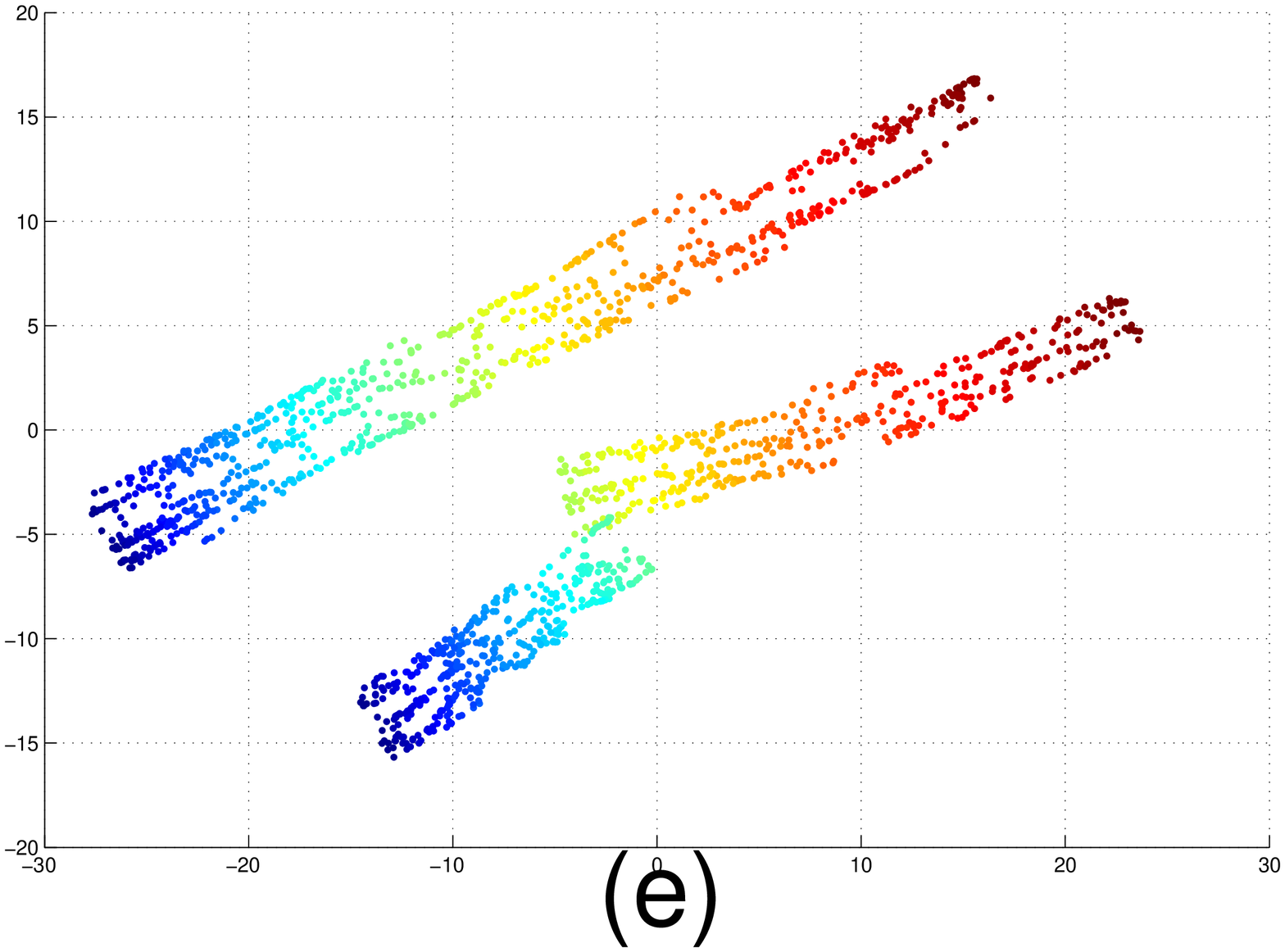}
\caption{Experiments on a three-manifolds data set. (a) The original
data set. (b) The result obtained by the k-CC Isomap. (c) The result obtained by the M-Isomap.
(d) Illustration of the procedure of the revised D-C Isomap.
(e) The result obtained by the revised D-C Isomap.
}\label{fig5}
\end{figure*}

In this subsection, we compare the k-CC Isomap, the M-Isomap and the revised
D-C Isomap on three 3-D data sets. It should be noted that in all experiments,
the neighborhood size is chosen corresponding to the best performance of each algorithm.

Fig. \ref{fig3} (a) is a two-manifolds data set with $N=1200$ data points, and
the data set is generated by the following matlab code:

{\ttfamily
\begin{tabular*}{8.5cm}{ll}
 &t=(1*pi/6)*(1+2*rand(1,N));\\
 &xx=t.*cos(t);yy=t.*sin(t);\\
 &zz  =[unifrnd(1,10,1,N/2)\quad unifrnd(16,25,1,N/2)];\\
 &X=[xx;zz;yy];\\
\end{tabular*}
}

It can be seen that each data manifold is intrinsically a
rectangular region with 600 data points. Fig. \ref{fig3}(b) shows the result
obtained by the k-CC Isomap, whose neighborhood graph is constructed by using
the $8$-CC method. It can be seen that the embedding shrinks along the
edges in the low-dimensional space and the edges of the embedding become noisy.
Fig. \ref{fig3}(c) shows the result obtained by the M-Isomap method with the
neighborhood size $k=8$. As can be seen, each data manifold is exactly unrolled,
and the inter-manifolds distance is precisely preserved. Fig. \ref{fig3}(d) illustrates
the initialization step of the revised D-C Isomap algorithm.
First, the two data manifolds $X^1$ and $X^2$ are identified.
Then the third data cluster $X^3$ is constructed, where the parameter $\lambda=0.1$.
Finally, the centers $O_1$ and $O_2$ of the data manifolds are computed by referring to
the nearest neighbors. The center of $X^3$ is also the data point $X^3$.
Fig. \ref{fig3}(e) shows the result of the revised D-C Isomap method.
It is seen that the embedding exactly preserves both the intra-manifold distances and
inter-manifolds distances.

Fig. \ref{fig4}(a) is another two-manifolds data set with $N=1200$ data points,
and the data set is generated by the following matlab code:

{\ttfamily
\begin{tabular*}{8.5cm}{ll}
 & t=[unifrnd(pi*11/12,pi*14/12,1,N/2)\\
 & \qquad unifrnd(pi*16/12,pi*19/12,1,N/2)];\\
 & xx=t.*cos(tt);yy=t.*sin(tt);\\
 & zz=unifrnd(1,25,1,N);\\
 & Y = [xx;zz;yy]; \\
\end{tabular*}
\\ \ \\
}
Each data manifold has $600$ data points. One data manifold is a
rectangular region and the other one is a round region. Fig. \ref{fig4}(b) shows the result
obtained by the k-CC Isomap with the neighborhood size $k=10$.
It can be seen that the rectangular region bent outwards and
the round region is prolonged. Fig. \ref{fig4}(c) shows the result obtained by the M-Isomap
method with the neighborhood size $k=8$. As can be seen, each data manifold is exactly
unrolled, and the inter-manifolds relationship is precisely preserved.
Fig. \ref{fig4}(d) illustrates the initialization step of the revised D-C Isomap algorithm.
The parameter $\lambda=0.5$ for the production of the new cluster $X^3$.
Fig. \ref{fig4}(e) shows the result of the revised D-C Isomap method with the
neighborhood size $k=5$. It can be seen that the embedding exactly preserves both the
intra-manifold distances and inter-manifolds distances.

Fig. \ref{fig5}(a) shows a three-manifolds data set with $N=1600$ data
points on the Swiss roll manifold. The data set is generated by the following matlab code:

{\ttfamily
\begin{tabular*}{8.5cm}{ll}
 & t1 = [unifrnd(pi*5/6,pi*16/12,1,N/4)];\\
 & t2 = [unifrnd(pi*18/12,pi*12/6,1,N/4)];\\
 & t3=(5*pi/6)*(1+7/5*rand(1,N/2));\\
 & a1=t1.*cos(t1); b1=t1.*sin(t1); \\
 &  c1=[unifrnd(-1,3,1,N/4)];\\
 & a2=t2.*cos(t2); b2=t2.*sin(t2); \\
 & c2=[unifrnd(-1,3,1,N/4)];\\
 & a3=t3.*cos(t3); b3=t3.*sin(t3); \\
 & c3=[unifrnd(6,10,1,N/2)];\\
 & x1=[a1;c1;b1];   x2=[a2;c2;b2];   x3=[a3;c3;b3]\\
 & Z=[x1\quad x2\quad x3]; \\
\end{tabular*}
\\ \ \\
}
There are three rectangular regions on the Swiss roll manifold.
The longest data manifold has $800$ data points, and each of the other two shorter
data manifolds contains $400$ data points. Fig. \ref{fig5}(b) shows the result
obtained by the k-CC Isomap algorithm with the neighborhood size $k=10$.
Due to the bad approximation of the inter-manifolds geodesics, edges of the
data manifolds bend outwards. Fig. \ref{fig5}(c) shows the result obtained by the
M-Isomap method, where the neighborhood size $k$ is set to be $8$.
As can be seen, all data manifolds are exactly unrolled, and the inter-manifolds
relationships of the three data manifolds are precisely preserved.
Fig. \ref{fig5}(d) illustrates the initiation step of the revised D-C Isomap algorithm.
The result of the revised D-C Isomap method is presented in Fig. \ref{fig5}(e).
As seen in Fig. \ref{fig5}(e), the embedding does not exactly preserve the
inter-manifolds distances. This is because the shape of the data manifolds are very
narrow. The selected reference data points can not efficiently relocate each piece of
the data manifold.

\subsection{Real world data sets}

\begin{figure*}
\includegraphics[width=5.5cm]{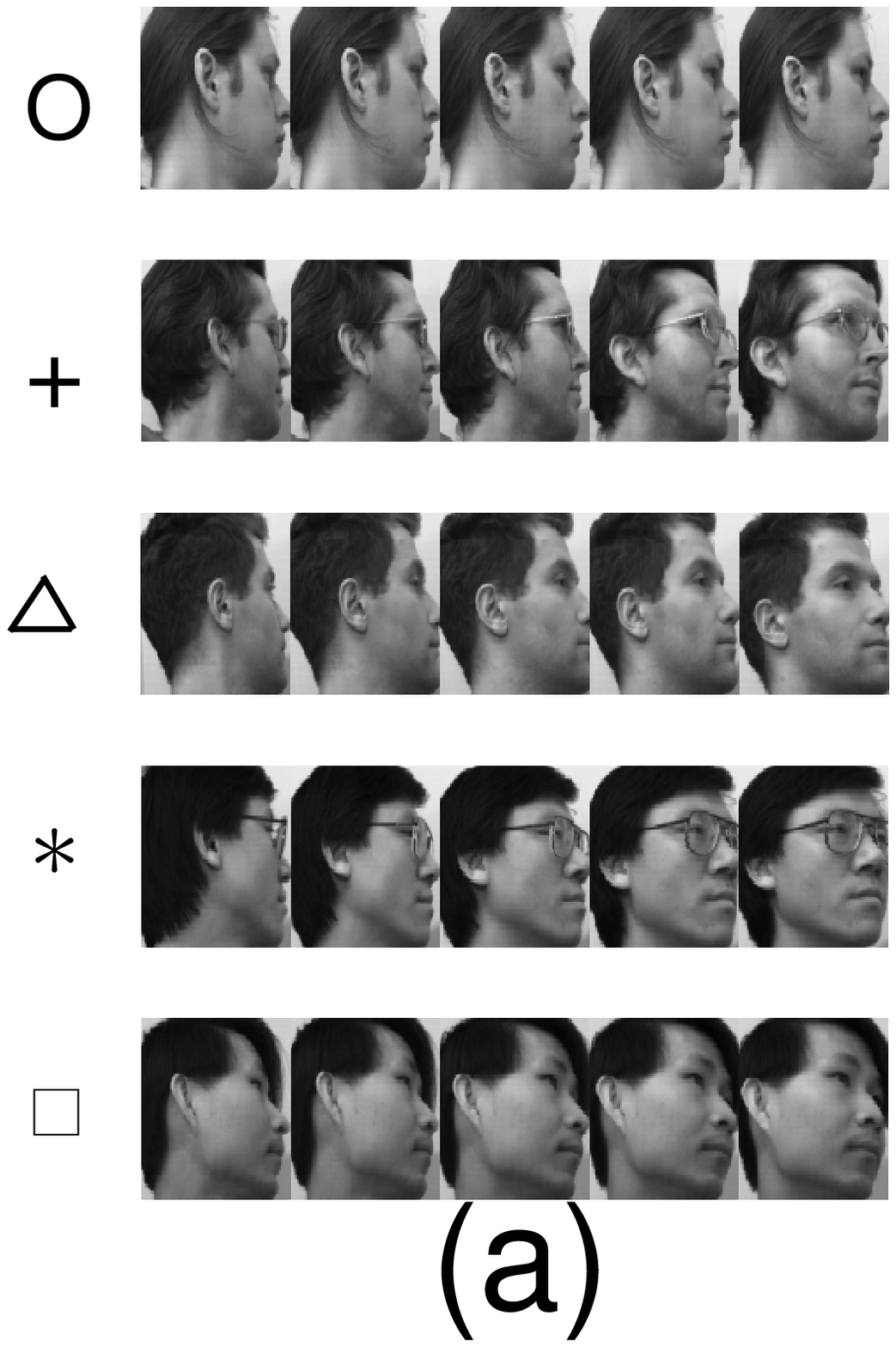}
\includegraphics[width=5.5cm]{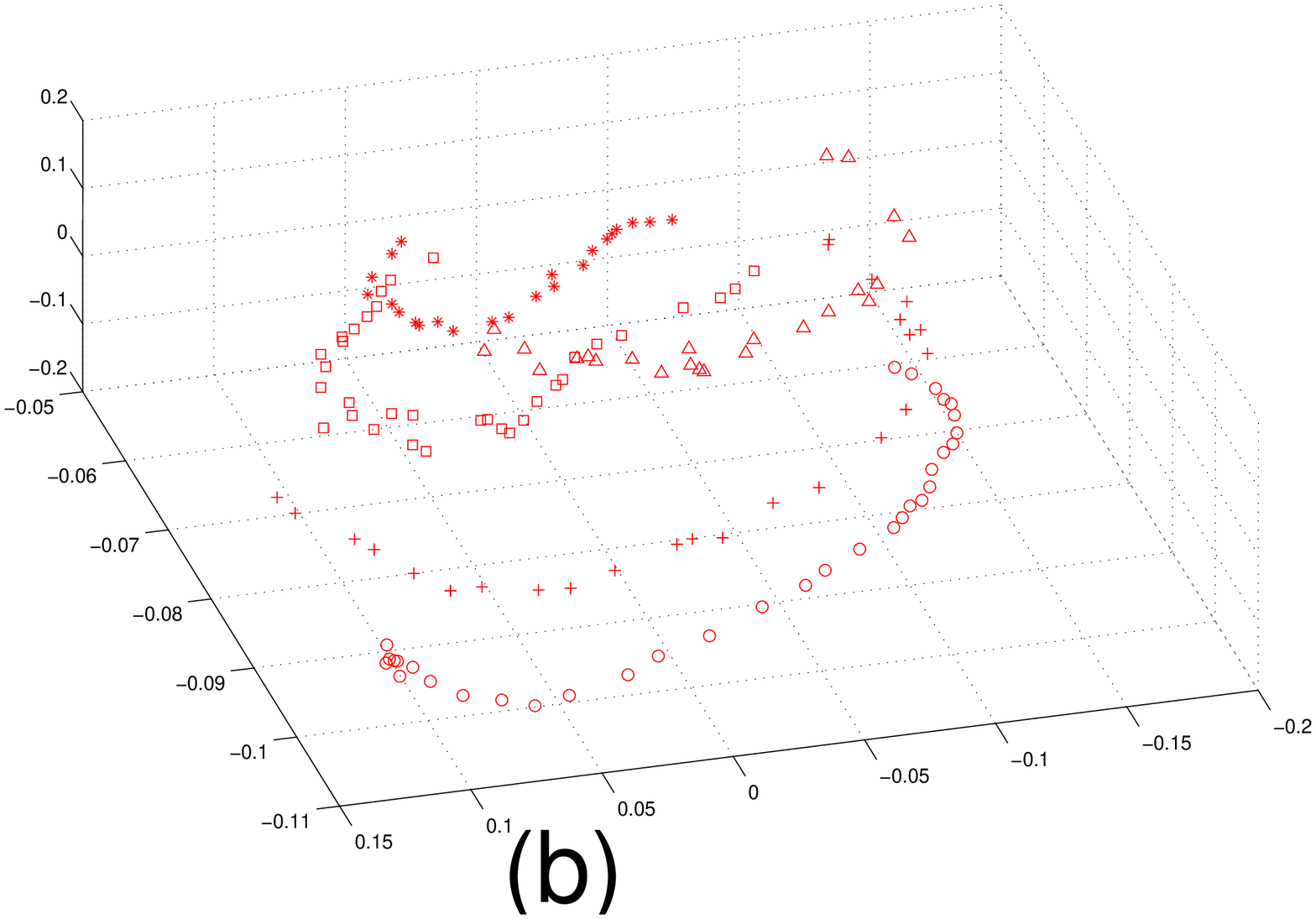}
\includegraphics[width=5.5cm]{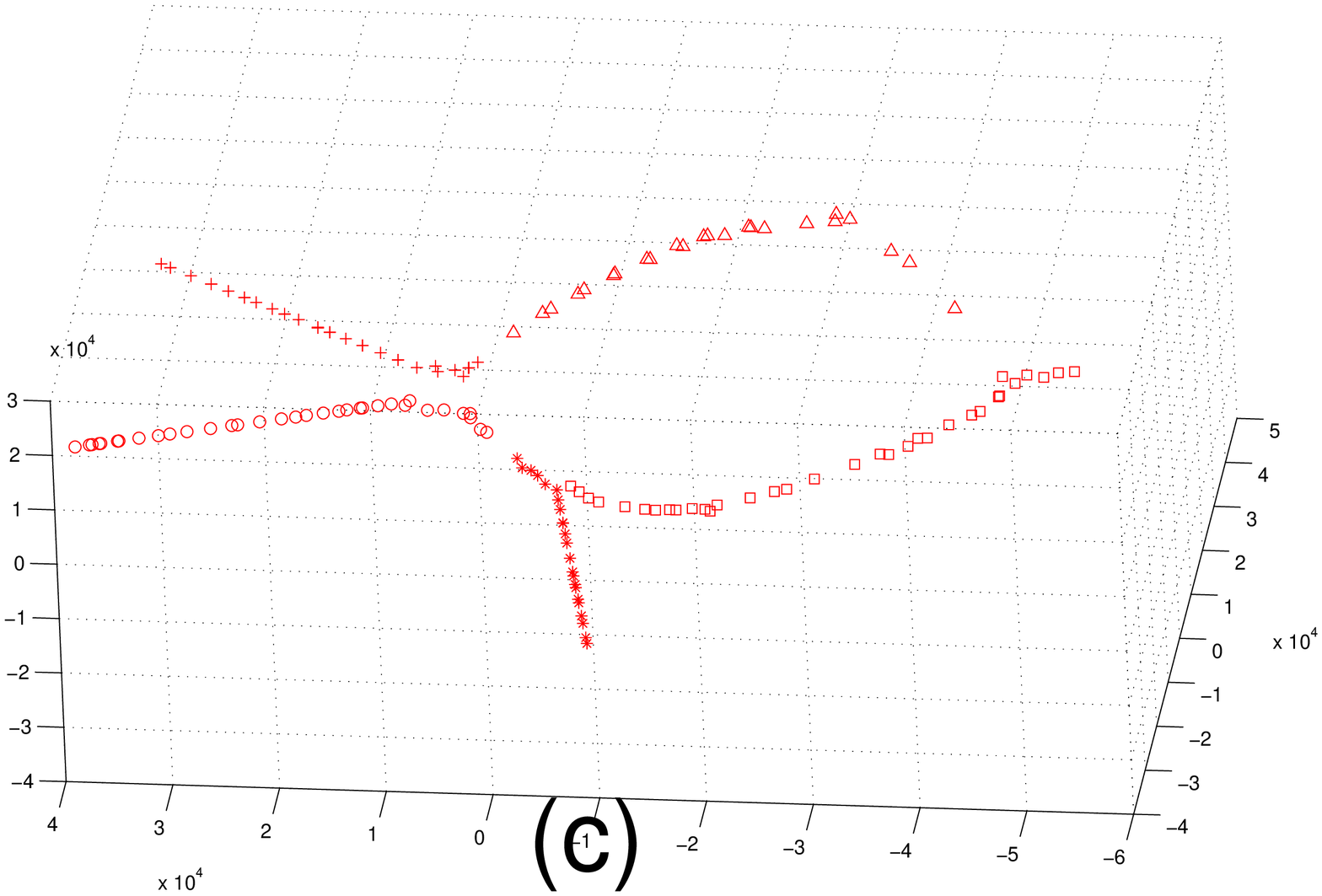}
\includegraphics[width=5.5cm]{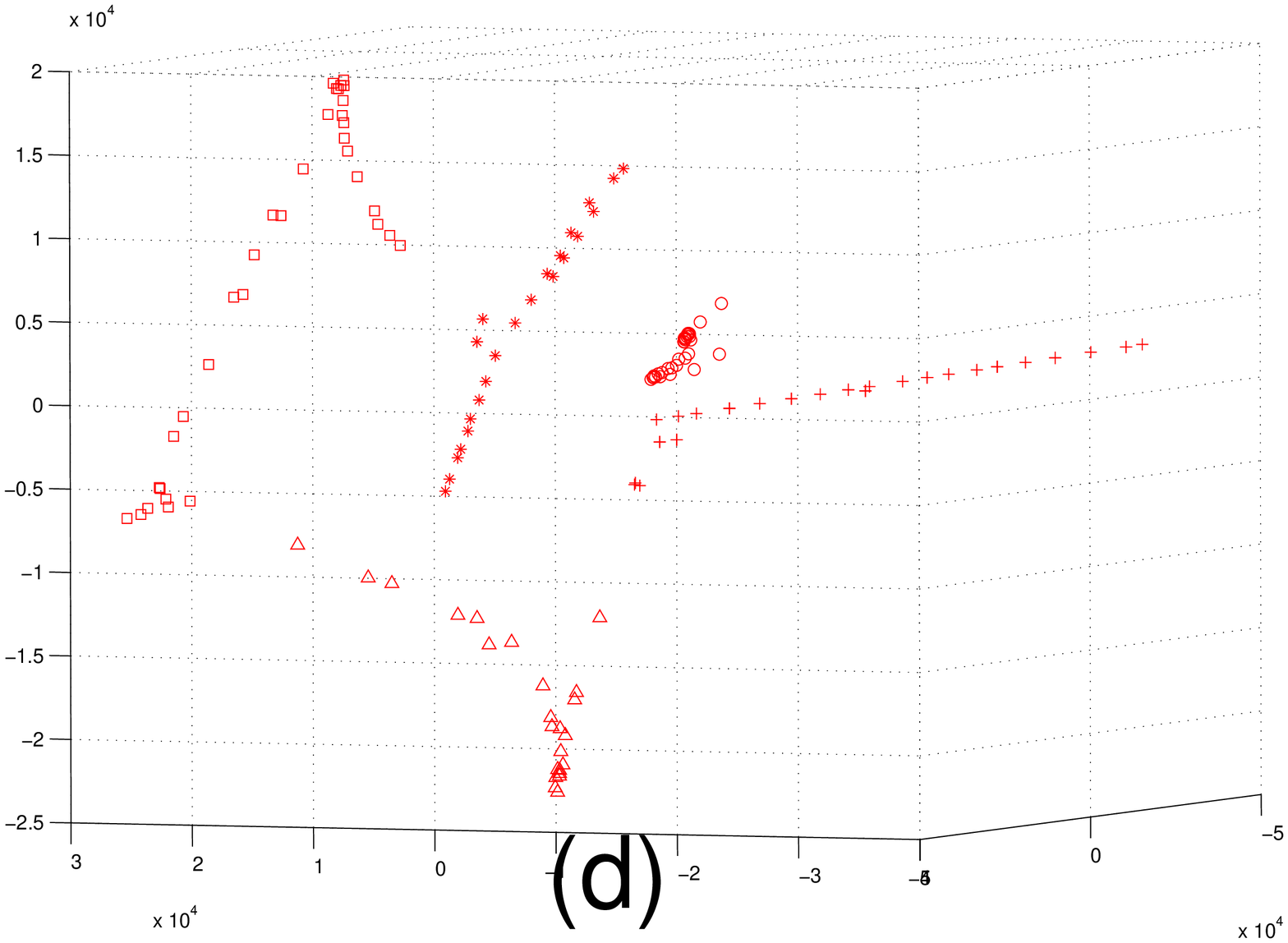}
\includegraphics[width=5.5cm]{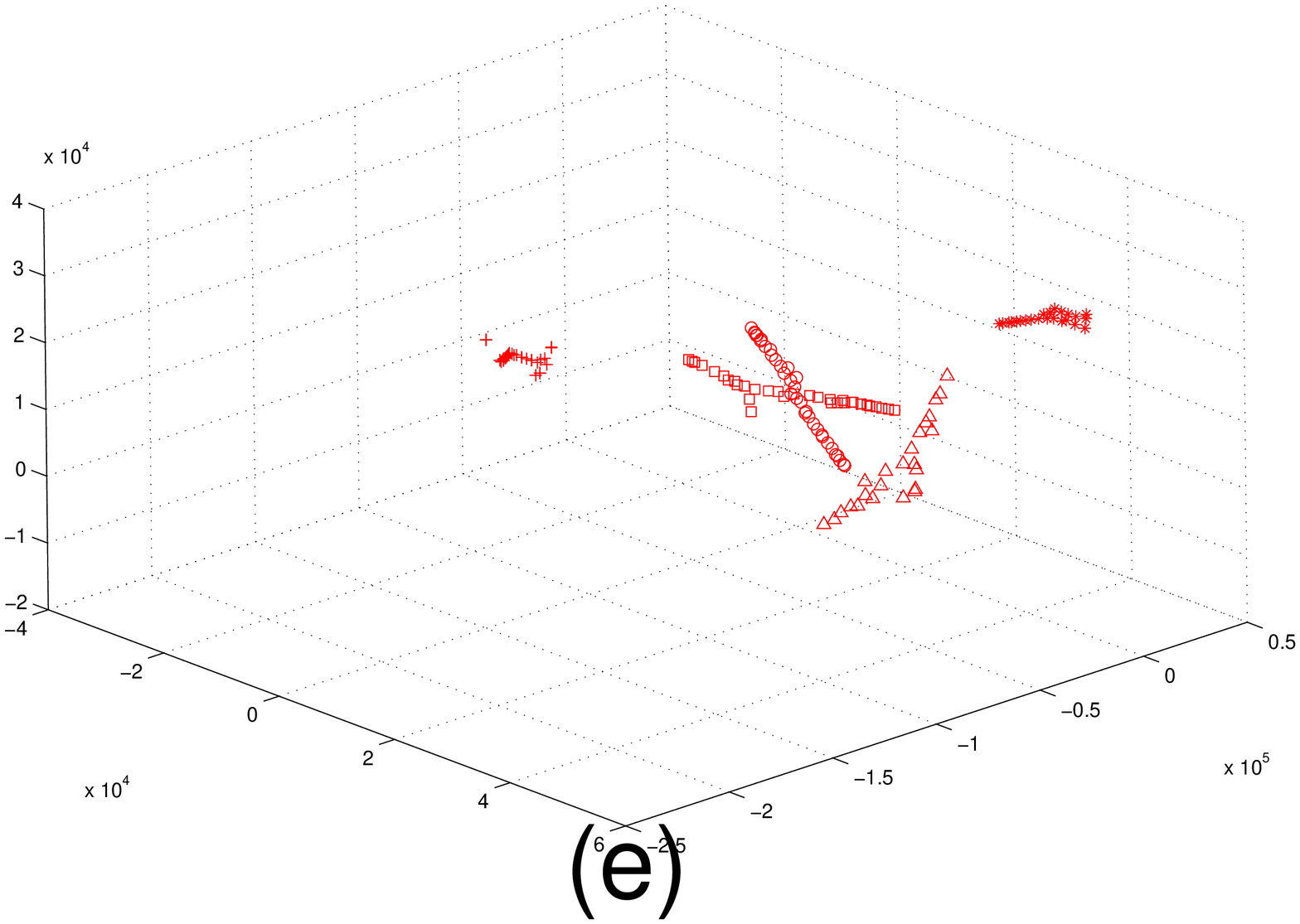}
\caption{(a) The face data set of five persons. (b) The result by PCA.
(c) The result by the k-CC Isomap. (d) The result by the M-Isomap. (e)
The result by the original D-C Isomap.
} \label{fig6}
\end{figure*}

\begin{figure*}
\begin{center}
\includegraphics[width=4cm]{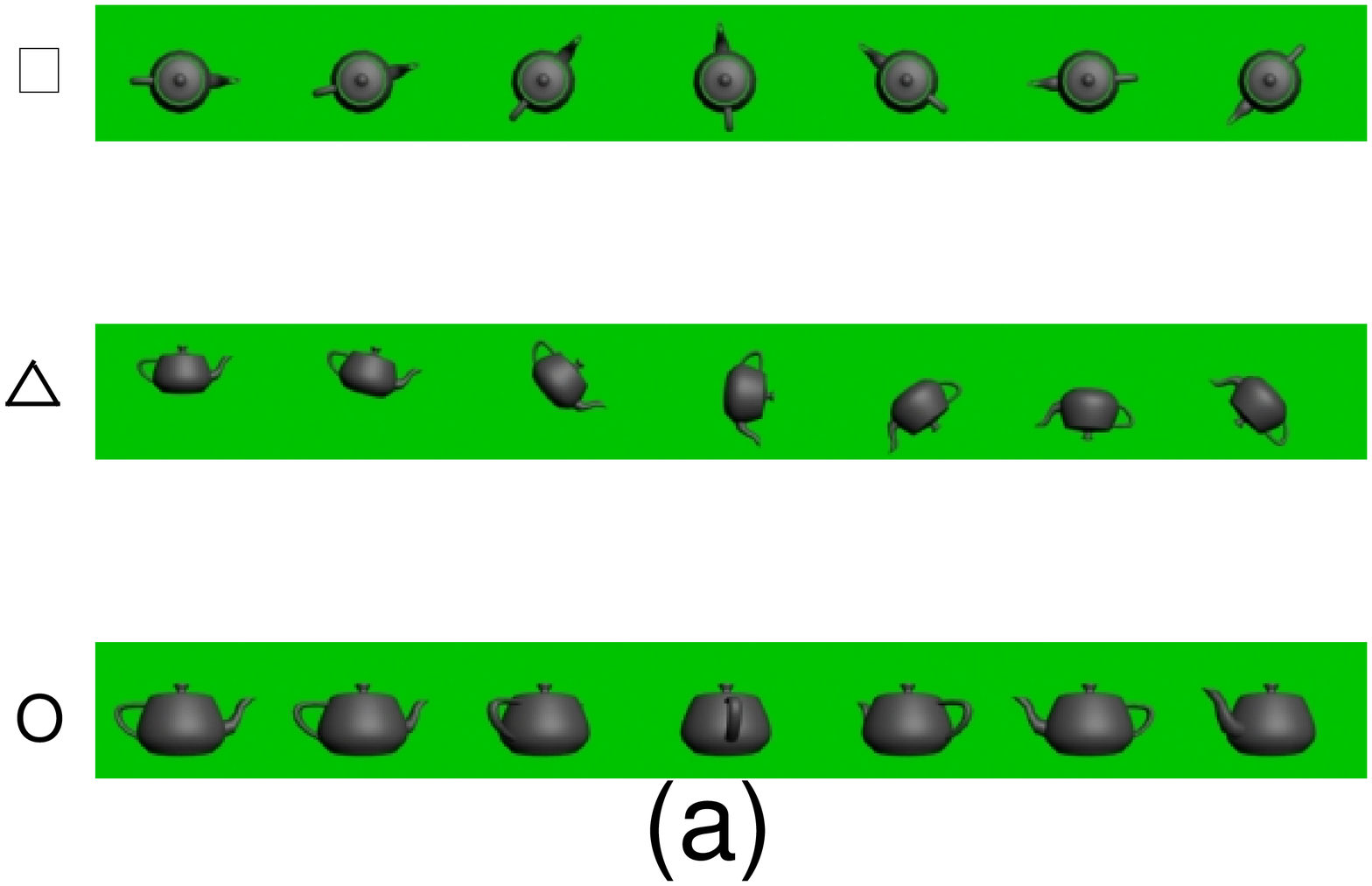}
\includegraphics[width=4cm]{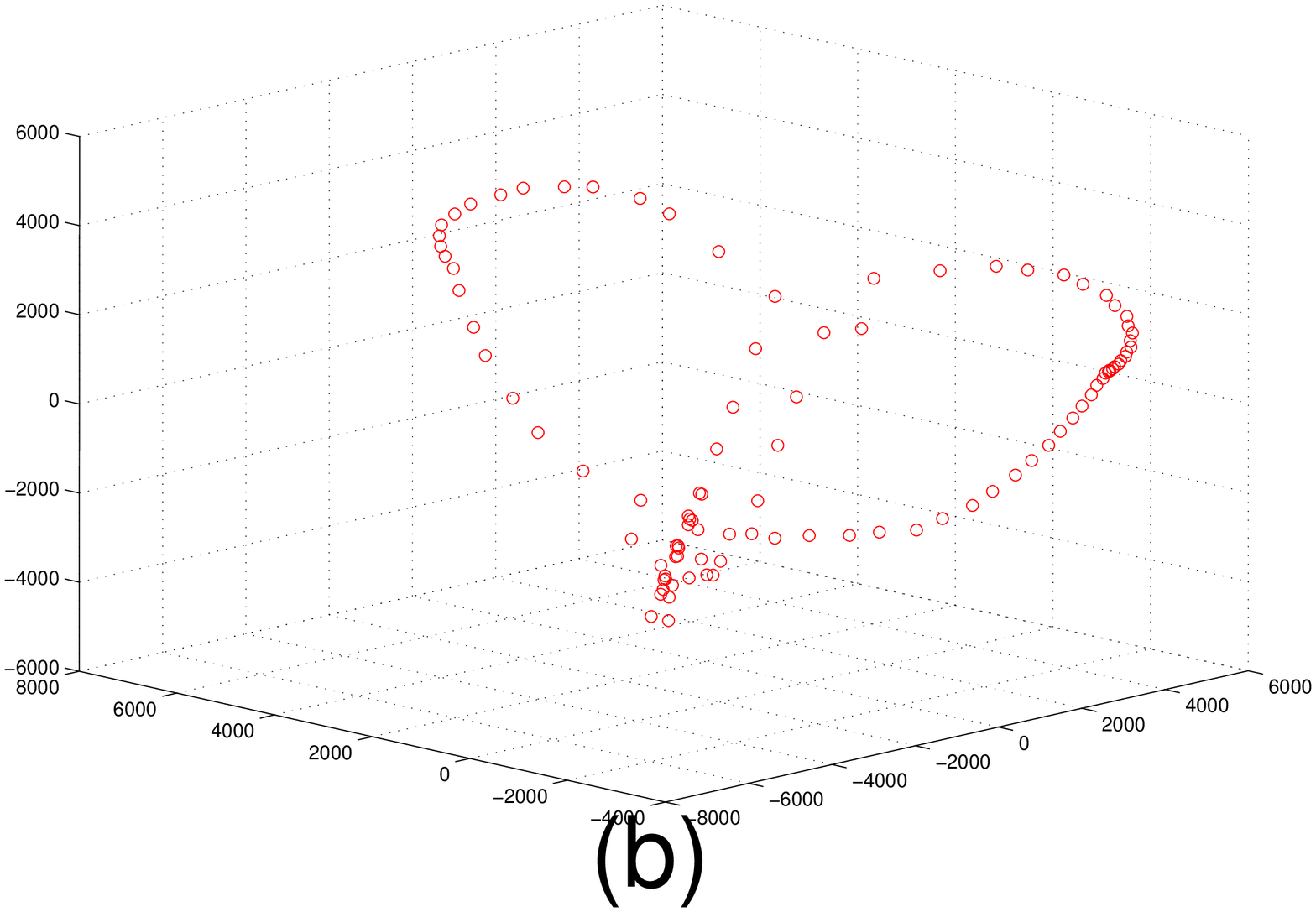}
\includegraphics[width=4cm]{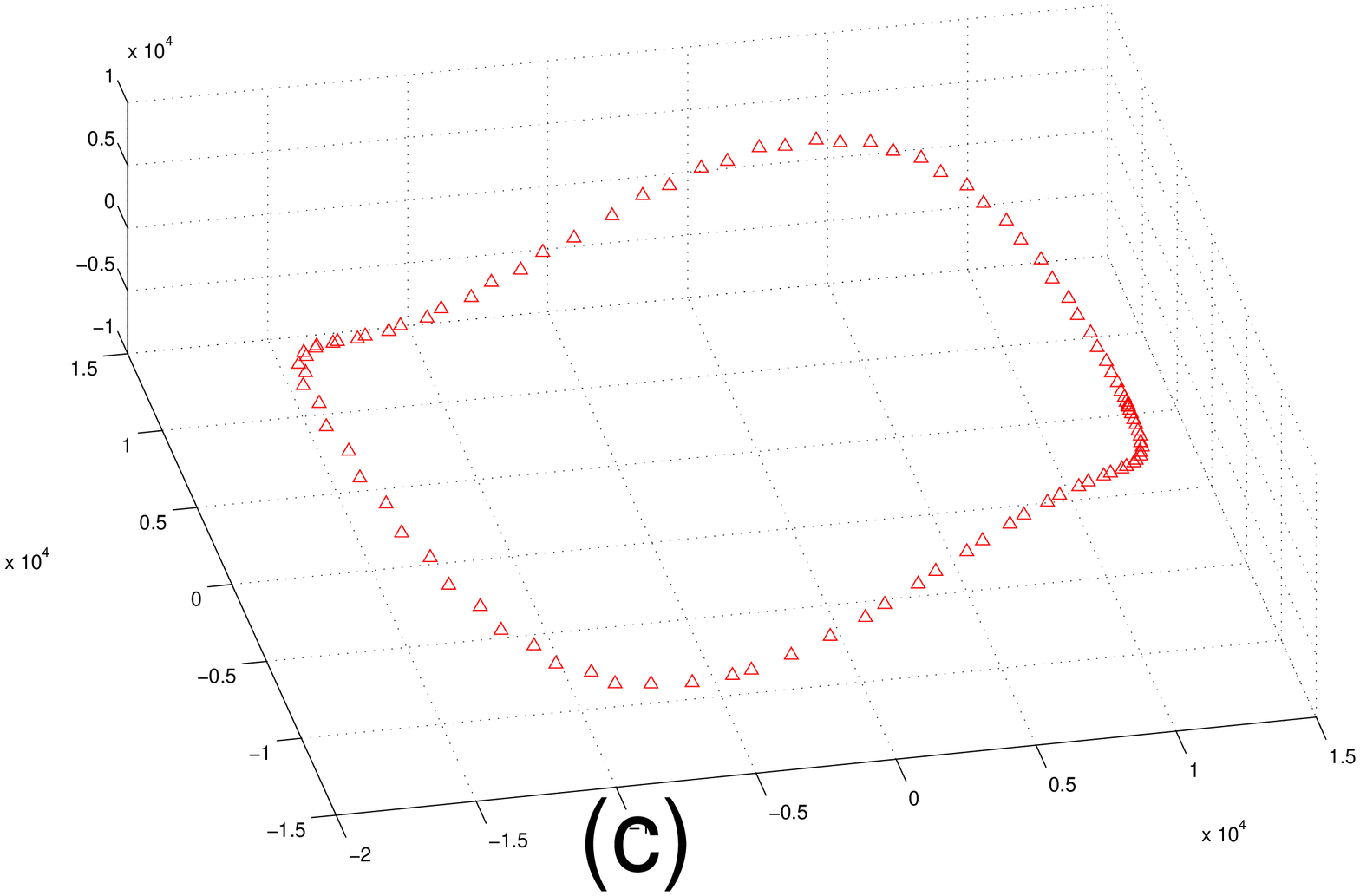}
\includegraphics[width=4cm]{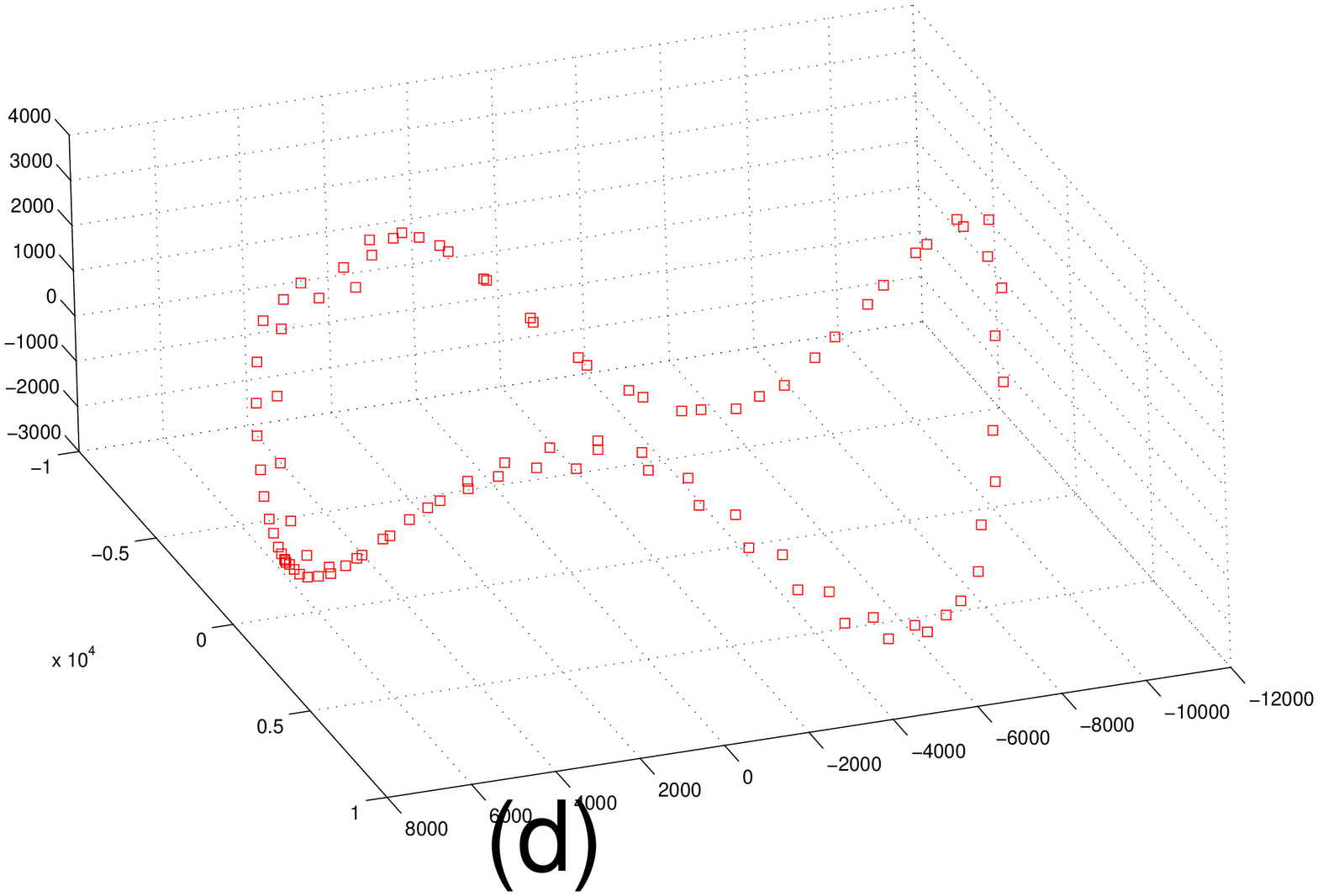}
\end{center}
\includegraphics[width=4cm]{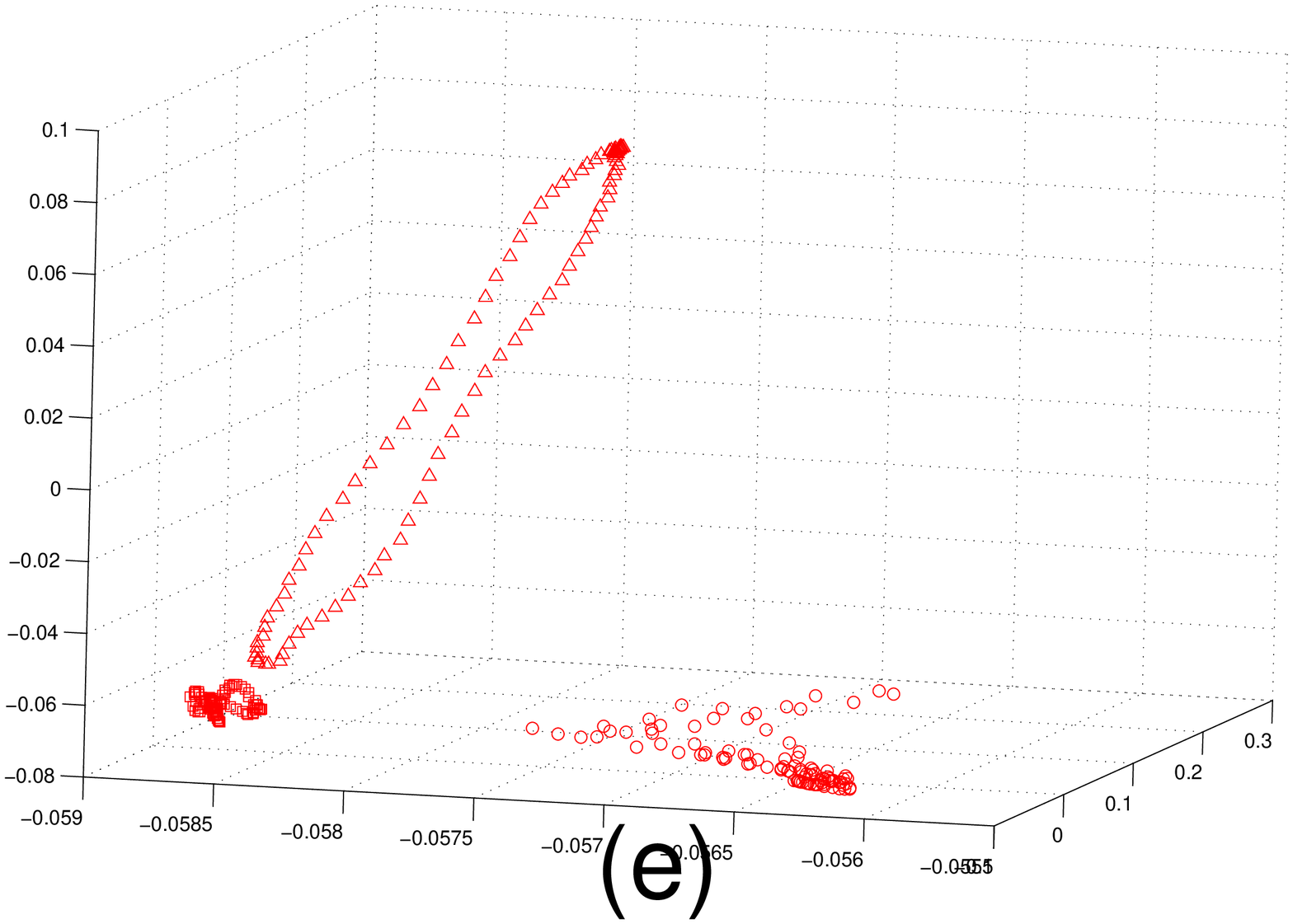}
\includegraphics[width=4cm]{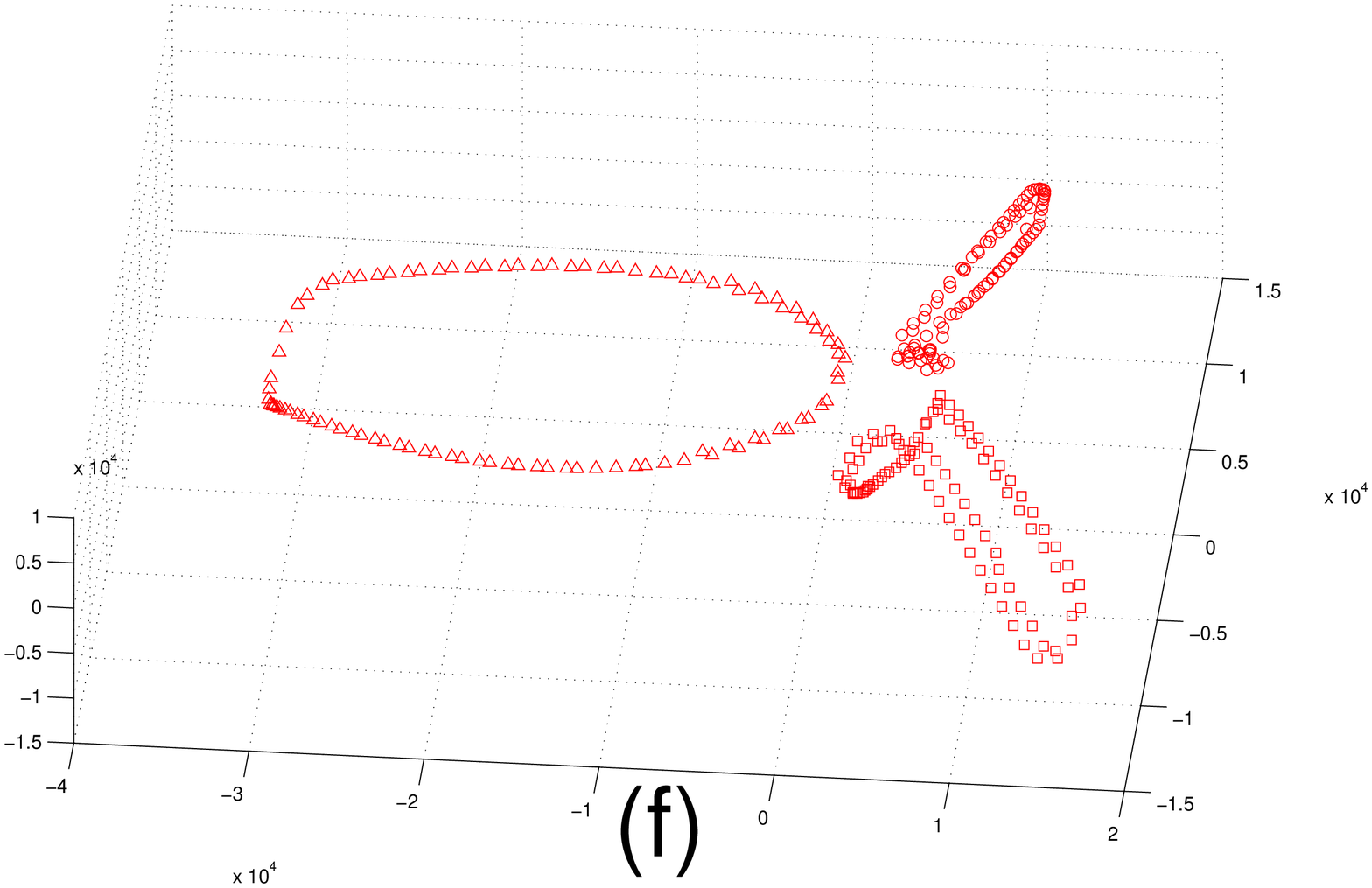}
\includegraphics[width=4cm]{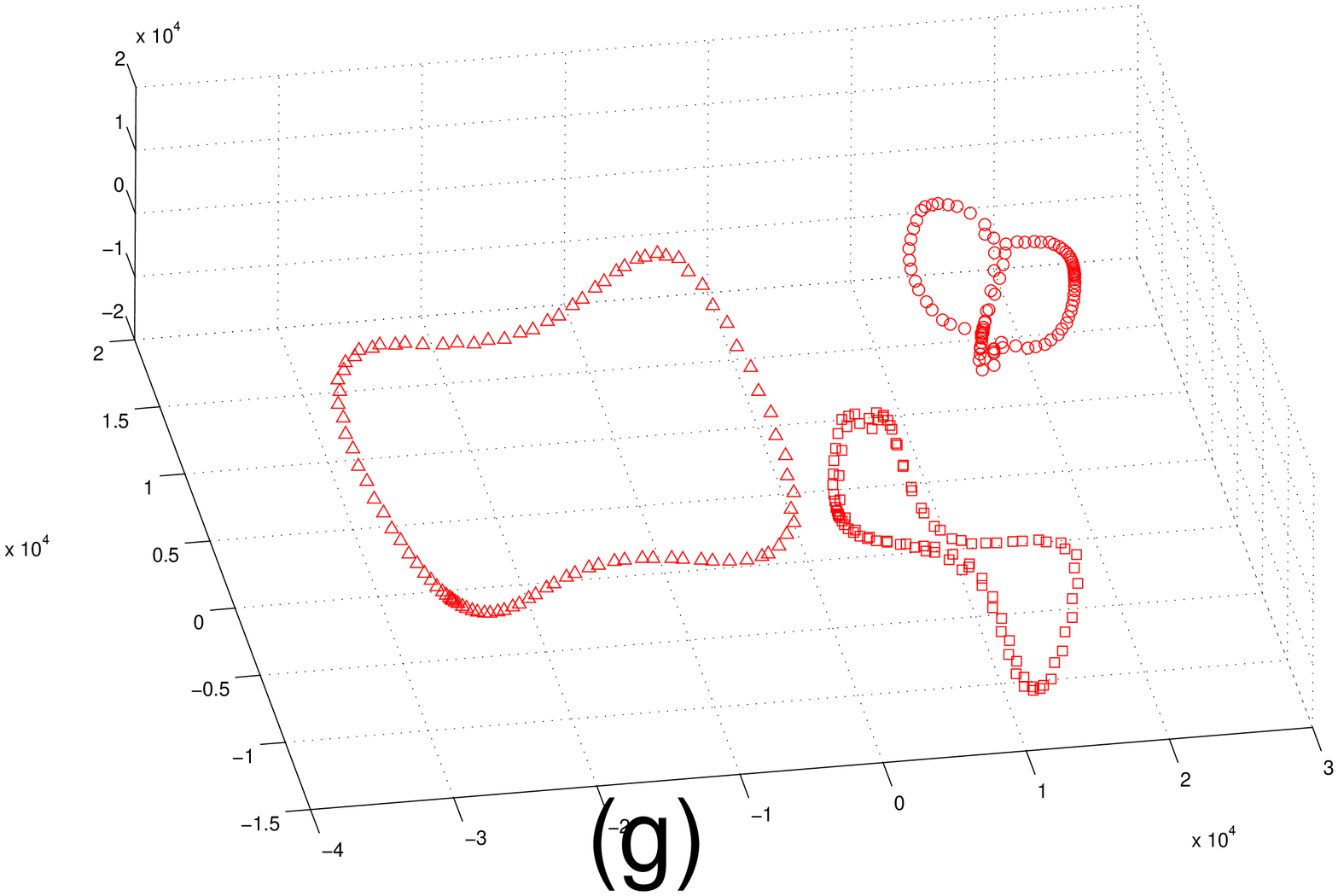}
\includegraphics[width=4cm]{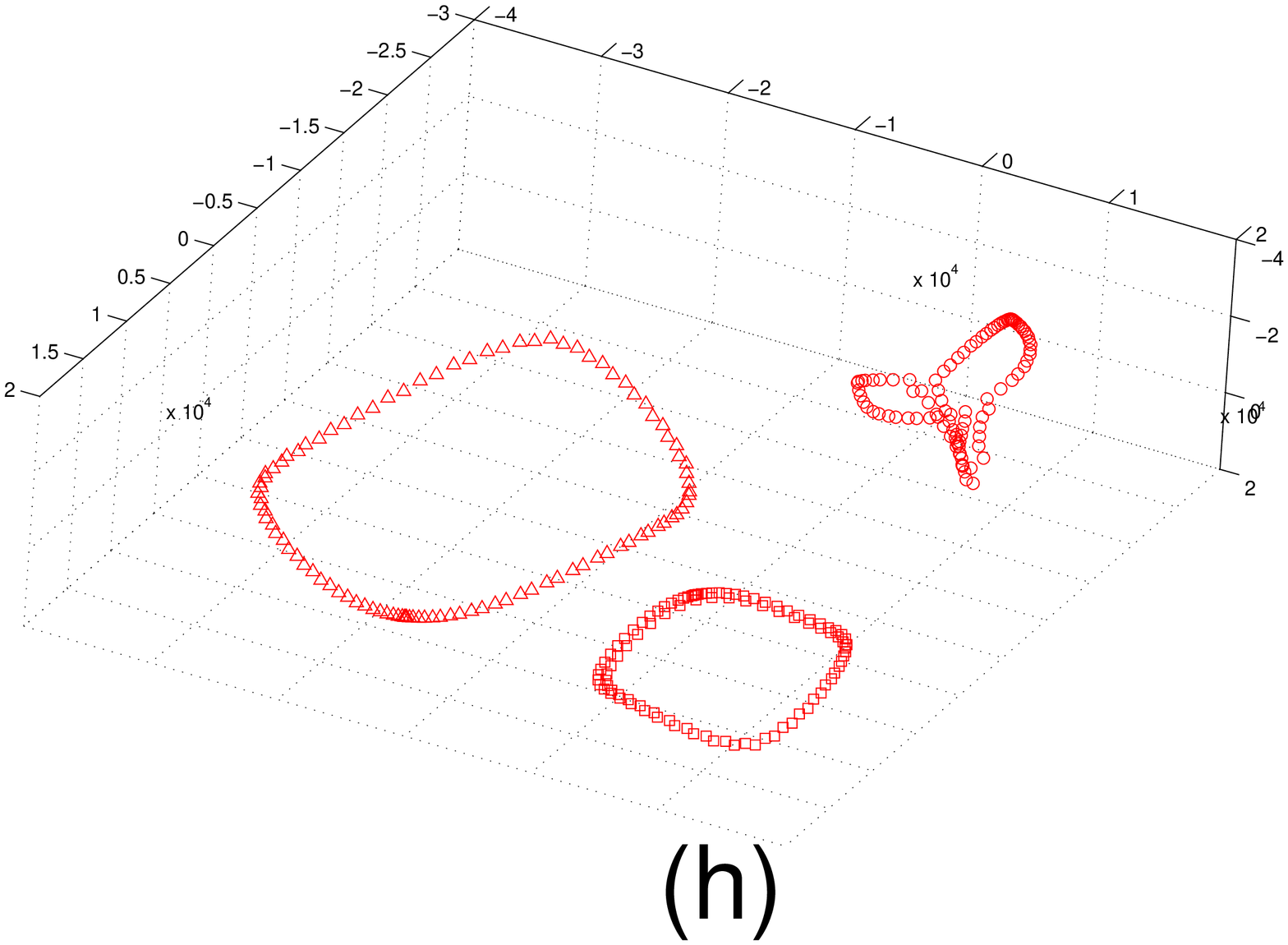}
\caption{(a) The teapot data set. In the experiments, '$\oblong$' stands
for the vertical view rotation of the teapot, '$\bigtriangleup $' stands for
the side view back-forth rotation of the teapot and '$\bigcirc $' stand for the
side view rotation of the teapot. (b) The result of Isomap on the teapot
vertical view rotation set. (c) The result of Isomap on the teapot
side view back-forth rotation set. (d) The result of Isomap on the
teapot side view rotation set. (e) The result of PCA on the teapot data
set. (f) The result of the k-CC Isomap on the teapot data set. (g) The
result of the M-Isomap on the teapot data set. (h) The result of the revised D-C Isomap
on the teapot data set.
}\label{fig7}
\end{figure*}

\begin{figure*}
\includegraphics[width=5.5cm]{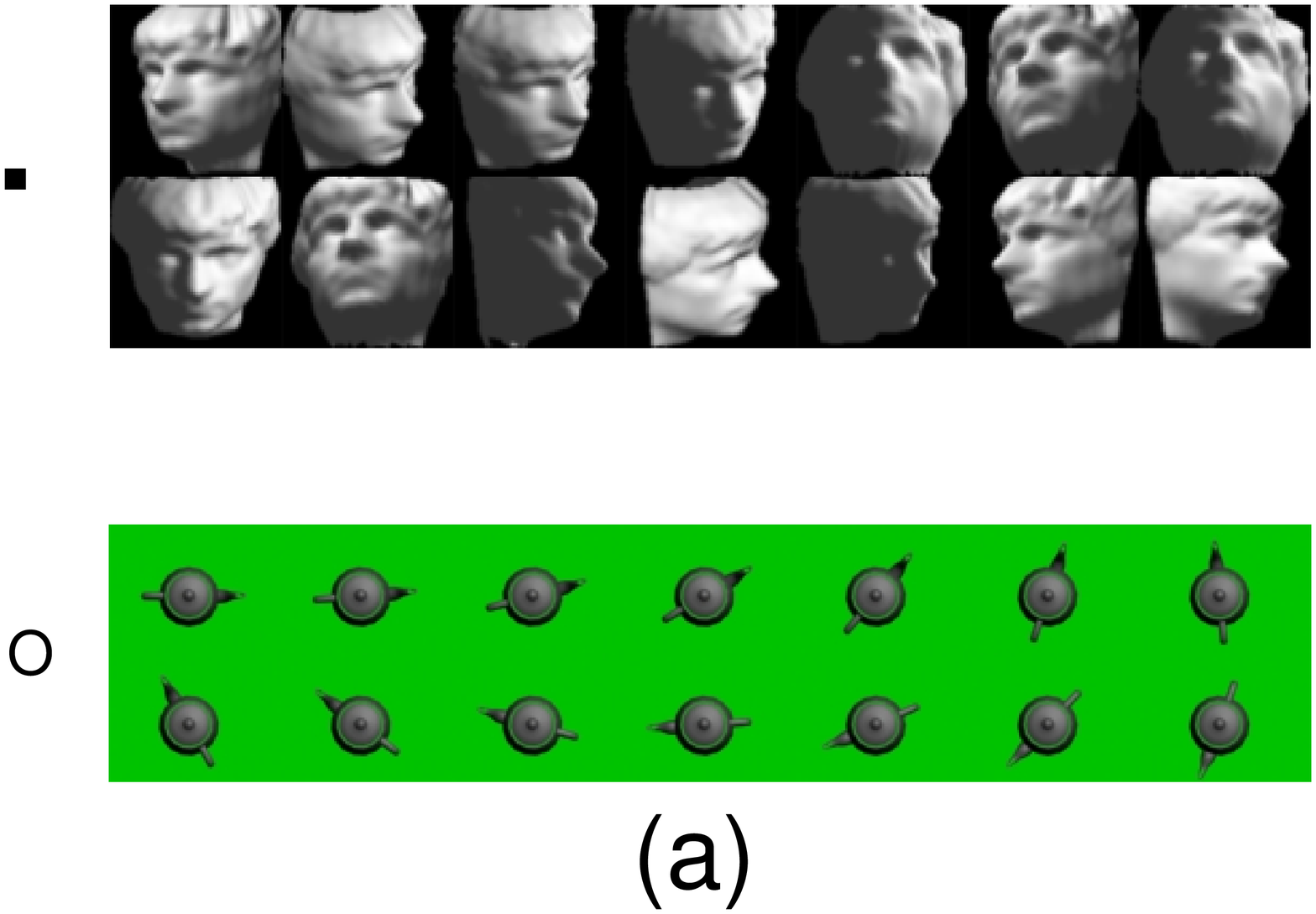}
\includegraphics[width=5.5cm]{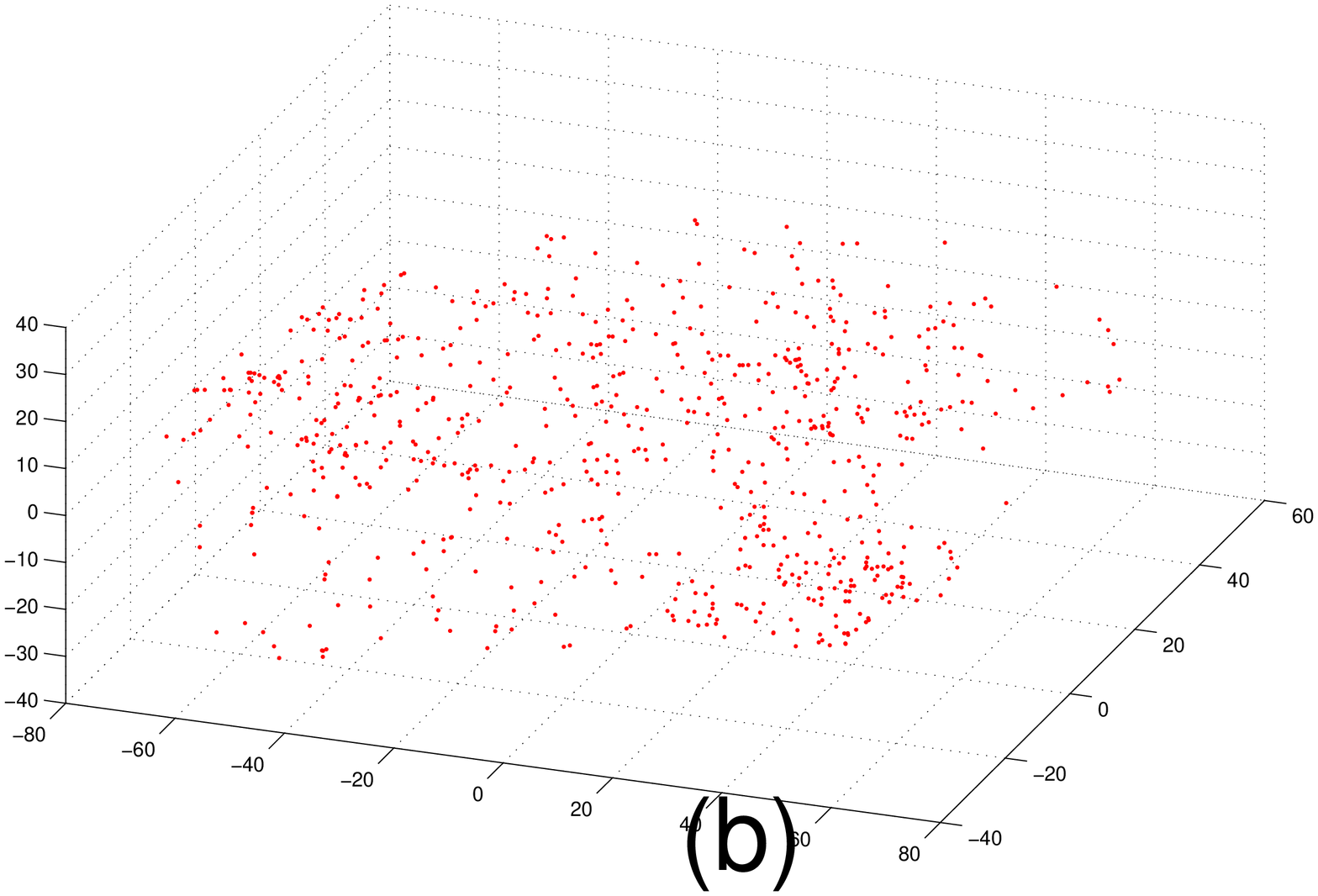}
\includegraphics[width=5.5cm]{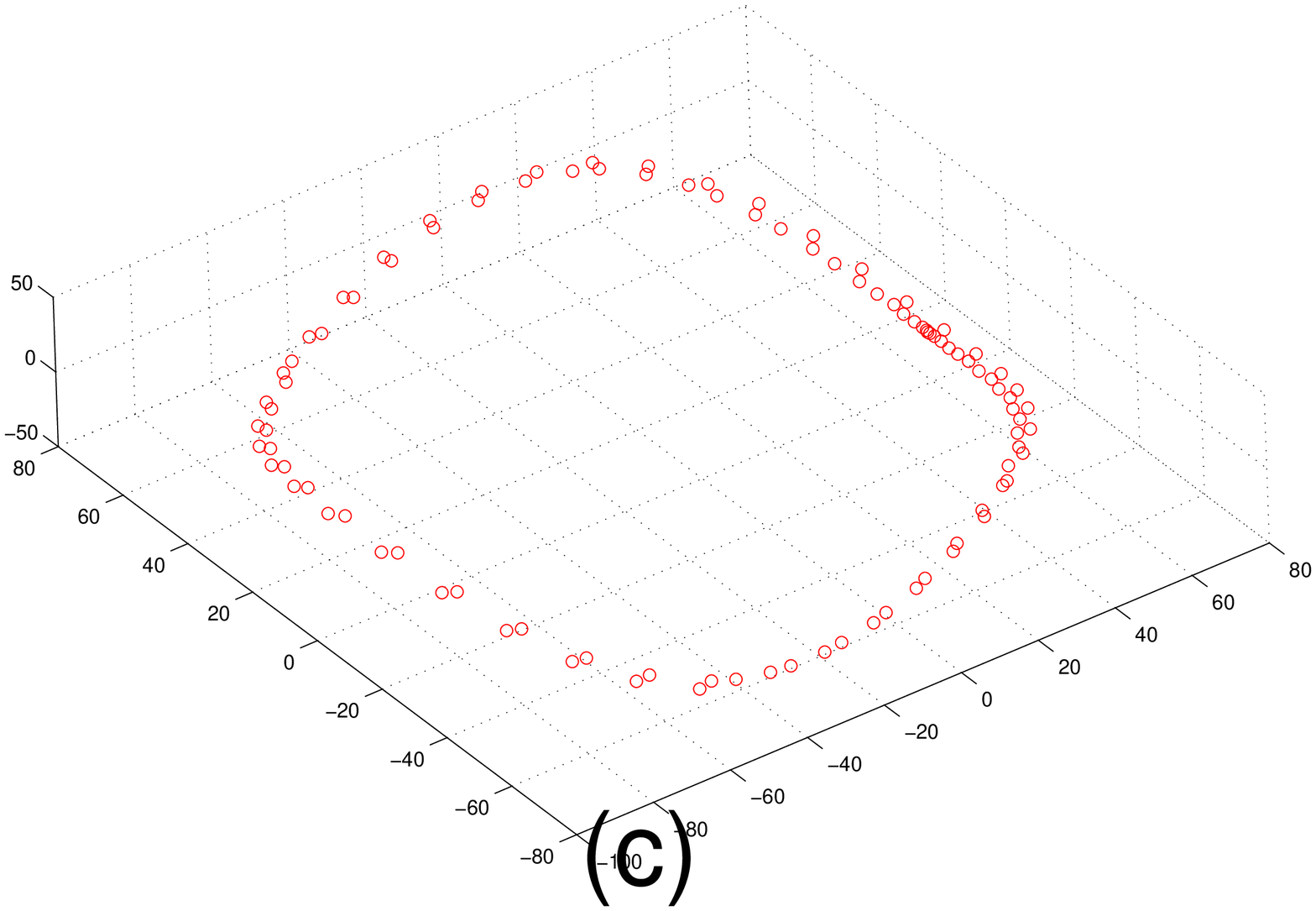}
\includegraphics[width=5.5cm]{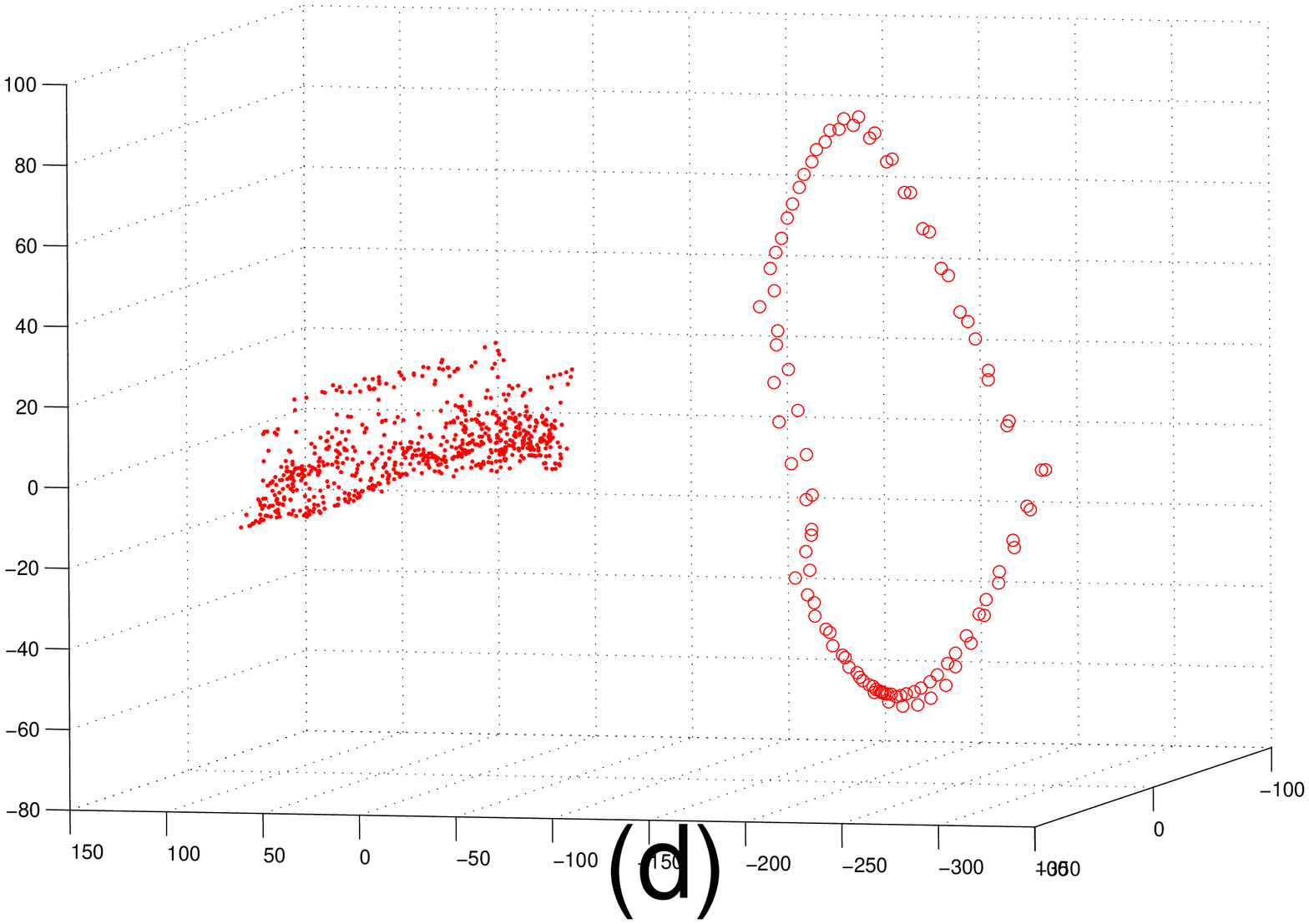}
\includegraphics[width=5.5cm]{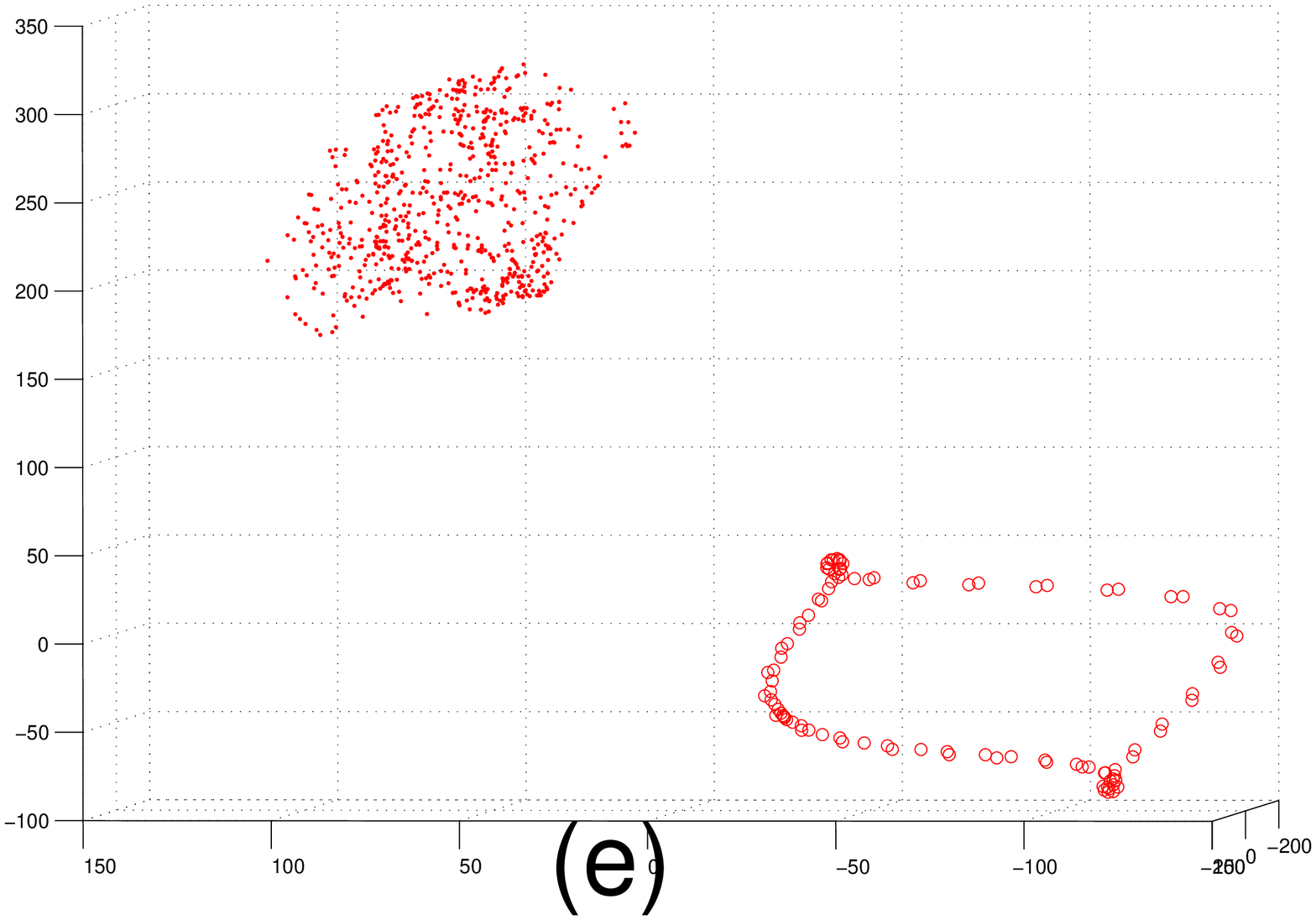}
\includegraphics[width=5.5cm]{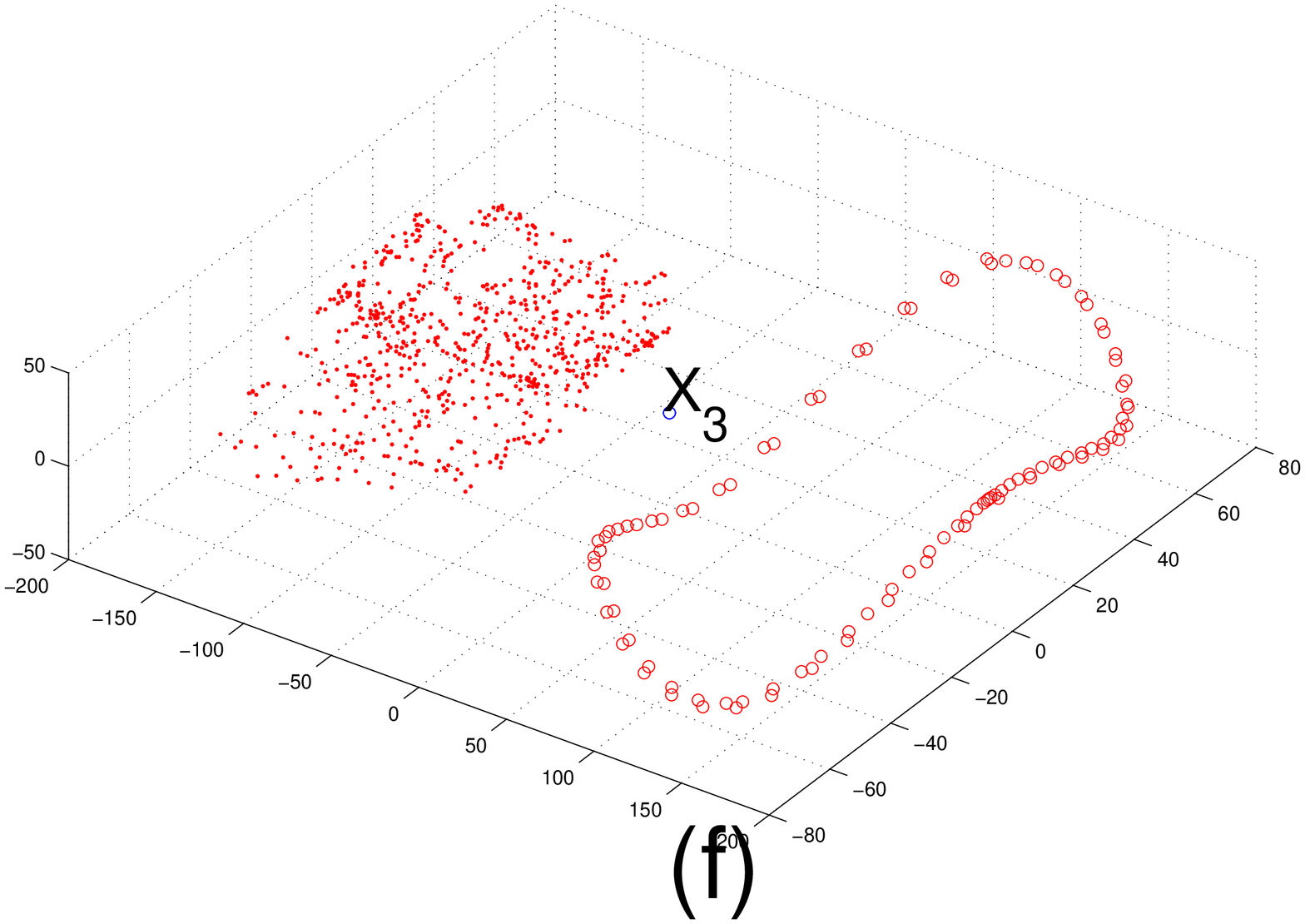}
\caption{(a) The IsoFACE and teapot data set, where '$\bullet $' stands for the IsoFACE data set
and '$\bigcirc $' stands for the teapot vertical view data set. (b) The result of
Isomap on the IsoFACE data set. (c) The result of Isomap on the teapot vertical view data set.
(d) The result of the k-CC Isomap on the IsoFACE and teapot data set.
(e) The result of the M-Isomap on the IsoFACE and teapot data set.
(f) The result of the revised D-C Isomap on the IsoFACE and teapot data set.
}\label{fig8}
\end{figure*}

Fig. \ref{fig6}(a) shows samples of the faces data \cite{faces5set} which contains
face images of five persons\footnote{http://www.cs.toronto.edu/~roweis/data.html}.
The data set consists of $153$ images and has $34$, $35$, $26$, $24$, $34$ images
corresponding to each face. These images are gray scale with resolution of $112\times92.$
They are transformed into vectors in a $10304$-dimensional Euclidean space.
In order to show the inter-manifolds relationship with more details,
the data is embedded into a three-dimensional space. Fig. \ref{fig6}(b) is the three-dimensional
embedding by the PCA method. It can be observed that data manifolds of faces are mixed up,
and the intra-face information is not preserved.
Fig. \ref{fig6}(c) is the result obtained by the k-CC Isomap method with $k=3$.
As seen from Fig. \ref{fig6}(c), although the data points are clustered, their inter-face
distances are not well preserved. The five lines are mixed up at one of their endings.
Fig. \ref{fig6}(d) shows the result obtained by the M-Isomap method with $k=3$.
Due to the limitation of the k-NN method in clustering, only two data manifolds are identified.
Although the data set is not well clustered, the result of the M-Ismap shows that the
low-dimensional embedding can be separated easily. Fig. \ref{fig6}(e) presents the result
obtained by the original D-C Isomap method, where the faces are split up beforehand.
The circumcenters are used as their centers. However, as can be seen, two faces are mixed up.

Fig. \ref{fig7}(a) presents samples of the teapot data set with $300$ data points,
where '$\boxempty $' stands for the teapot bird-view images, '$\Delta $' stands for the
teapot back-forth rotation images and '$\bigcirc $' stands for the teapot side-view images.
Each image is an $80\times60\times3$ RGB colored picture, that is, a vector in a $14400$-dimensional
input space. The data points do not distribute on a single global manifold, which
is a great challenge to the classical manifold learning methods.
The experiments show that the three data manifolds can be identified by the k-CC method.
In order to show their exact embedding, Fig. \ref{fig7}(b)-(d) present the embedding of
each data manifold by the classical Isomap with the neighborhood size $k=3$.
Fig. \ref{fig7}(e) gives the result obtained by the PCA method.
It can be seen that the data set is clustered but the shape of each embedding is deformed
because of the linear characteristic of the PCA method. Fig. \ref{fig7}(f) is the result
obtained by the k-CC Isomap method with the neighborhood size $k=3$.
The bad approximations of the inter-manifolds geodesics lead to the deformation of the embedding
in the low-dimensional space. Fig. \ref{fig7}(g) shows the result obtained by the M-Isomap
method with the neighborhood size $k=3$. From Fig. \ref{fig7}(g), it is seen clearly that
the data set is clearly clustered and the intra-manifolds relationships are exactly preserved.
Fig. \ref{fig7}(h) gives the result obtained by the revised D-C Isomap method with
the neighborhood size $k=3$. It is seen that the revised D-C Isomap algorithm also
produces a satisfactory result.

Fig. \ref{fig8}(a) shows samples of the IsoFACE and teapot rotation bird-view
data set. The IsoFACE data set consists of $698$ images and each image is a
$64\times64$ ($4096$-dimensional) gray scale picture. Since the input dimension of
the IsoFACE data set is different from that of the teapot data set,
the dimension of the IsoFACE set is increased by adding zeros to the bottom of the face image vectors.
The scale of the teapot data set should also be changed such that the scales of the two embeddings
is compatible. The teapot data vectors are divided by $100$, that is, the scale of
the teapot data points shrinks to $\frac{1}{100}$ of the original one.
Fig. \ref{fig8}(b) is the 3-D embedding of the IsoFACE data set by the classical Isomap
with the neighborhood size $k=5.$ Fig. \ref{fig8}(c) is the 3-D embedding of the scaled teapot data set
obtained by the classical Isomap with the neighborhood size $k=5$.
Fig. \ref{fig8}(d) gives the result obtained by the $5$-CC Isomap method.
It can be seen that the shape of the IsoFACE data set is distorted badly.
Fig. \ref{fig8}(e) shows the result obtained by the M-Isomap method with the neighborhood size $k=5.$
The performance of the M-Isomap method is significantly improved compared with that of the k-CC Isomap.
Fig. \ref{fig8}(f) presents the result obtained by the revised D-C Isomap method, which is also
satisfactory.

\subsection{Discussion}

In our experiments, there are several important features which should be considered:

\begin{enumerate}
\item Since the k-CC Isomap tries to preserve poor and good approximations of
geodesics simultaneously, its low-dimensional embedding is usually deformed.
This method works well if each data manifold has comparable number of data points
and the data manifolds can not be very far from each other.
The algorithm does not work well otherwise.

\item The revised D-C Isomap overcomes some limitations of the original D-C Isomap.
Meanwhile, the robustness of the algorithm is also enhanced by adding a new fictitious cluster.

\item The M-Isomap connects data manifolds with multiple edges, which can control
the rotation of the low-dimensional embedding, and at the same time,
better preserves inter-manifolds distance. Similarly to the D-C Isomap algorithm,
the M-Isomap can also isometrically preserve intra-manifold geodesics and inter-manifolds distances.
\end{enumerate}

To sum up, Table \ref{tb2} shows the comparison of the general performance of the five versions of
Isomap algorithms: classical Isomap, k-CC Isomap , Original D-C Isomap, revised D-C Isomap
and M-Isomap. The labels "$\Delta$" stands for poor performance, "$ \boxempty $"
stands for not bad and "$\bigcirc$" stands for good.
"Density" means the generalization ability on manifolds with different density, that is,
different neighborhood sizes, "Dimensionality" means the generalization ability on manifolds
with different intrinsic dimensionality, "Isometric" means the property of isometry in
preserving the inter and intra-manifold relationship and "Generalization" means
the overall generalization ability to learn from multiple data manifolds.

\begin{table}
\caption{The generalization performance of classical Isomap, k-CC Isomap,
Original D-C Isomap, revised D-C Isomap and M-Isomap for multi-manifolds learning.
}\label{tb2}
\begin{center}
\begin{tabular}{lllll}
\hline
Method & Density & Dimensionality &  Isometric & Generalization \\
\hline
classical & $\Delta $ & $\Delta $ & $\Delta $& $\Delta$\\
k-CC  & $\bigcirc$ & $\bigcirc$ & $\Delta $ & $ \boxempty $\\
Original D-C & $\bigcirc$ & $\bigcirc$ & $\bigcirc$ & $ \boxempty $ \\
revised D-C & $\bigcirc$ & $\bigcirc$ & $\bigcirc$ & $ \bigcirc $  \\
M-Isomap & $\bigcirc$& $\bigcirc$& $\bigcirc$& $\bigcirc$\\
\hline
\end{tabular}
\end{center}
\end{table}

\section{Conclusion}\label{sec6}

In this paper, the problem of multi-manifolds learning is discussed.
A general procedure for isometric multi-manifolds learning is proposed.
The procedure can be used to build multi-manifolds learning algorithms
which are able to faithfully preserve not only intra-manifold geodesic distances
but also the inter-manifolds geodesic distances.
The M-Isomap is an implementation of the procedure and shows promising results
in multi-manifolds learning. Compared with the k-CC Isomap which was also introduced in this paper
based on the general procedure, the M-Isomap has the advantage of low computational complexity.
With the procedure, the D-C Isomap proposed in \cite{DCIsomap} was revised to overcome
some of its limitations. Compared with the original D-C Isomap, the revised D-C Isomap
is more effective in learning multi-manifolds data sets.
Experiments have also been conducted on both synthetic and images data sets to illustrate the
efficiency of the above multi-manifolds learning algorithms.
Future work will be conducted on the application of the multi-manifolds learning algorithms.

\section*{Acknowledgment}

The work of the second author (HQ) was partly supported by the
Chinese Academy of Sciences through the Hundred Talents Program, the
NNSF of China under grant 60675039 and grant 60621001, the 863
Program of China under grant 2006AA04Z217 and the Outstanding Youth
Fund of the NNSF of China under grant 60725310. The third author
(BZ) was partly supported by the Chinese Academy of Sciences through
the Hundred Talents Program, the 863 Program of China under grant
2007AA04Z228, the 973 Program of China under grant 2007CB311002 and
the NNSF of China under grant 90820007.



\begin{thebibliography}{1}

\bibitem{PCA}I. T. Jolliffe, ``Principal Component Analysis,''
\emph{Springer-Varlag,} New York, 1989. 

\bibitem{MDS} T. F. Cox and M. A. Cox, ``Multidimensional Scaling,''
\emph{Chapman \& Hall,} London, 2001. 

\bibitem{ScienceMachLearning}E. Mjolsness and D. De Coste, ``Machine learning for science:
State of the art and future prospects,'' \emph{Science,} vol. 293, pp. 2051-2055, Sep. 2001.

\bibitem{ISOMAP2000} J.B. Tenenbaum, V. de Sliva and J. C. Landford,
``A global geometric framework for nonlinear dimensionality reduction,''
\emph{Science,} vol. 290, pp. 2319-2323, Dec. 2000.

\bibitem{LLE2000} S.T. Roweis and L. K. Saul, ``Nonlinear dimensionality reduction by
local linear embedding,'' \emph{Science,} vol. 290, pp. 2323-2326, Dec. 2000.

\bibitem{manifoldPerception} H.S. Seung and D.D. Lee, ``The manifold ways of perception,''
\emph{Science,} vol. 290, pp. 2268-2269, Dec. 2000.

\bibitem{laplacianEigenmap}M. Belkin and P. Niyogi, ``Laplacian eigenmaps for dimensionality
reduction and data representation,'' \emph{Neural Computation,} vol. 15, no 6, pp. 1373-1396, June 2003.

\bibitem{hessianLLE}D. Donoho and C. Grimes, ``Hessian eigenmaps: new locally linear
embedding techniques for high-dimensional data,'' \emph{Proc. Nat. Acad. Sci. USA},
vol. 100, no. 10, pp. 5591-5596, 2003.

\bibitem{LTSA} Z. Zhang and H. Zha, ``Principal manifolds and nonlinear dimensionality
reduction via tangent space alignment,'' \emph{SIAM J. Sci. Comput.}, vol. 26, pp. 313-338, 2004.

\bibitem{DiffusionMaps} R. R. Coifman, S. Lafon, A.B. Lee, M. Maggoni, B. Nadler, F. Warner
and S.W. Zuck, ``Geometric diffusions as a tool for harmonic analysis and structure definition
of data: Diffusion maps,'' \emph{Proc. Nat. Acad. Sci. USA}, vol. 102, no. 21, pp. 7426-7431, May. 2005.

\bibitem{RML} T. Lin and H. Zha, ``Riemannian Manifold Learning,''
\emph{IEEE Trans. Pattern Analysis and Machine Intelligence,} vol. 30, no. 5, pp. 796-809, May. 2008.

\bibitem{ContinuumIsomap} H. Zha and Z. Zhang, ``Continuum Isomap for manifold learning,''
\emph{Computational Statistics \& Data Analysis,} vol. 52, issue. 1, pp. 184-200, Sep. 2007.

\bibitem{incrementaIsomap} M.H.C. Law and A.K. Jain, ``Incremental nonlinear dimensionality
reduction by manifold learning,'' \emph{IEEE Trans. Pattern Analysis and Machine Intelligence},
vol. 28, no. 3, pp. 377-391, March. 2006.

\bibitem{proof} M. Bernstein, V. de Silva, J.C. Langford and J.B. Tenenbaum,
``Graph approximations to geodesics on embedded manifolds,'' Technical report,
Dept. of Psychology, Stanford Univ., Dec. 2000.

\bibitem{ClusteringSurvey} M. Filippone, F. Camastra, F. Masulli and S. Rovetta,
``A survey of kernel and spectral methods for clustering,'' \emph{Pattern Recognition,}
vol. 41, pp. 176-190, May. 2007.

\bibitem{Classify-Isomap} M.-H. Yang, ``Extended Isomap for pattern classification,''
\emph{ICPR 2002: Proc. Int. Conf. Pattern Recognition,} vol. 3, Aug. 2002. 

\bibitem{Classify-laplacian} X. He, S. Yan, T. Hu, P. Niyogi and H. Zhang,
``Face recognition using Laplacianfaces,'' \emph{IEEE Trans. Pattern Analysis
and Machine Intelligence,} vol. 27, no. 3, pp. 328-340, March. 2005.

\bibitem{Semi-learning} M. Belkin, V. Sindhwani and P. Niyogi, ``Manifold regularization: a
geometric framework for learning from examples,''
\emph{J. Machine Learning Research,} vol. 7, pp. 2399-2434, Dec. 2006.

\bibitem{NDRforVisual&Classify} X. Geng, D. Zhang and Z. Zhou, ``Supervised nonlinear
dimensionality reduction for visualization and classification,''
\emph{IEEE Trans. Systems, Man and Cybern. B,} vol. 35, no. 6, pp. 1098-1107, Dec 2005.

\bibitem{STisomap} O.C. Jenkins and M.J. Mataric, ``A Spatio-temporal extension to
Isomap nonlinear dimension reduction,'' \emph{ACM. Proc. 21st Int. Conf. on Machine learning,}
vol. 69, pp. 56-66, 2004.

\bibitem{IsomapTimeSeries} A. Rahimi, B. Recht and T. Darrell, ``Learning to transform time series
with a few examples ,'' \emph{IEEE Trans. Pattern Analysis and Machine Intelligence,}
vol. 29, no. 10, pp. 1759-1775, Oct. 2007.

\bibitem{IsomapTopoStability} J.B. Tenenbaum, V. de Sliva and J.C. Landford ``Response to
Comments on the Isomap algorithm and topological stability,''
\emph{Sciences,} vol. 295, no. 5552, pp. 7, Jan. 2002.

\bibitem{ExtendedIsomaplearning} Y. Wu and K.L. Chan ``An extended Isomap algorithm for learning
multi-class manifold,'' \emph{Proc. of the 2004 Int. Conf. on Machine Learning and Cybernetics,}
vol. 6, pp. 3429-3433, Aug. 2004.

\bibitem{DCIsomap} D. Meng, Y. Leung, T. Fung and Z. Xu ``Nonlinear dimensionality reduction
of data lying on the multicluster manifold,'' \emph{IEEE Trans. Systems, Man and Cybern. B,}
vol. 38, issue. 4, pp. 1111-1122. Aug. 2008.

\bibitem{CMVM} B. Li, D. Huang, C. Wang and K. Liu ``Feature extraction using constrained
maximum variance mapping,'' \emph{Pattern Recognition,} vol. 41, pp. 3287-3294. May. 2008.

\bibitem{kMST2004} L. Yang, ``K-Edge Connected Neighborhood Graph for Geodesic Distance
Estimation and Nonlinear Projection,'' \emph{Proc. of 17th Int. Conf. on Pattern Recognition (ICPR'04),}
vol. 1, pp. 196-199. 2004.

\bibitem{MinkST2005} L. Yang, ``Building k Edge-Disjoint Spanning Trees of Minimum Total
Length for Isometric Data Embedding,'' \emph{IEEE Trans. Pattern Analysis and Machine Intelligence,}
vol. 27, no. 10, pp. 1680-1683, Oct. 2005.

\bibitem{kEC2005} L. Yang, ``Building k Edge-connected neighborhood graph for distance-based
data projection,'' \emph{Pattern Recognition Letters,} vol. 26, no. 13, pp. 2015-2021, Oct. 2005.

\bibitem{KVC2006} L. Yang, ``Building k-Connected Neighborhood Graphs for Isometric
Data Embedding,'' \emph{IEEE Trans. Pattern Analysis and Machine Intelligence,}
vol. 28, issue. 5, pp. 827-831, May. 2006.

\bibitem{APSDsMST} D. Zhao and L. Yang, ``Incremental Construction of Neighborhood
Graphs for Nonlinear Dimensionality Reduction,''
\emph{Proc. of Int. Conf. on Pattern Recognition,} vol. 28, no. 5, pp. 827-831, May. 2006.

\bibitem{APSDs} D. Zhao and L. Yang, ``Incremental Isometric Embedding of High-Dimensional
Data Using Connected Neighborhood Graphs,'' \emph{IEEE Trans. Pattern Analysis and Machine Intelligence,}
vol. 31, no. 1, pp. 86-98, Jan. 2009.

\bibitem{SelectionParameterIsomap} O. Samko, A.D. Marshall and P.L. Rosin `` Selection of the optimal
parameter value for the Isomap algorithm,'' \emph{Pattern Recognition Letters},
vol. 27, no. 9, pp. 968-979, Feb. 2006.

\bibitem{ClusteringAlgorithms} R. Xu and D. Wunsch II, ''Survey of clustering algorithms,''
\emph{IEEE Trans on Neural Networks,} vol. 16, no. 3, pp. 645-678, May. 2005.

\bibitem{fractal2002} F. Camastra and A. Vinciarelli, ''Estimating the intrinsic dimension
of data with a fractal-based approach,'' \emph{IEEE Trans. Pattern Analysis and Machine
Intelligence} vol. 24, no. 10, pp. 1404-1407, Oct. 2002.

\bibitem{MLE2004} E. Levina and P. J. Bickel ``Maximum likelihood estimation of
intrinsic dimension,'' \emph{NIPS 04: Neural Information Processing Systems},2005.

\bibitem{revisedMLE} D.J.C. MacKay and Z. Ghahramani, ''Comments on 'Maximum likelihood
estimation of intrinsic dimension' by E. Levina and P. Bickel (2005),'' \emph{in:
http://www.inference.phy.cam.ac.uk/mackay/dimension/}

\bibitem{incisingball} M. Fan, H. Qiao and B. Zhang, ''Intrinsic dimension estimation
of manifolds by incising balls,'' \emph{Pattern Recognition} vol. 42, no. 5, pp. 780-787,
May. 2009.

\bibitem{faces5set} B. Graham and N.M. Allinson, ''Characterizing Virtual Eigensignatures
for General Purpose Face Recognition'', In: H. Wechsler, P.J. Phillips, V. Bruce, F.
Fogelman-Soulie and T.S. Huang (eds): ''Face Recognition: From Theory to Applications;''
NATO ASI Series F, \emph{Computer and Systems Sciences}, Vol. 163, pp. 446-456, 1998.

\bibitem{ManiCluster}D. Kushnir, M. Galun and A. Brandt ``Fast multiscale clustering and
manifold identification,'' \emph{Pattern Recognition}, vol. 28, no. 10, pp. 1876-1891, Oct. 2006.

\end{thebibliography}
\end{document}